\documentclass[review]{elsarticle}

\graphicspath{ {./figures/} }
\usepackage{hyperref}
\usepackage{float}
\usepackage{verbatim} 
\usepackage{subfig}
\usepackage{graphicx}
\usepackage{amsfonts}
\usepackage{algorithm2e}
\RestyleAlgo{ruled}
\usepackage{amsmath}
\restylefloat{figure}
\restylefloat{table}
\usepackage{bbm}
\usepackage{mathtools}
\usepackage{amssymb}
\usepackage{multirow}
\usepackage[table,xcdraw]{xcolor}

\usepackage[labelsep=period]{caption}

\newtheorem{definition}{Definition}

\newtheorem{lemma}{Lemma}
\newtheorem{proof}{Proof}

\journal{Computers in Industry}

\begin{document}
\begin{frontmatter}

\title{Transfer learning of state-based potential games for process optimization in decentralized manufacturing systems}

\author[label1]{Steve Yuwono\corref{cor1}}
\ead{yuwono.steve@fh-swf.de}

\author[label2]{Dorothea Schwung}
\ead{dorothea.schwung@hs-duesseldorf.de}

\author[label1]{Andreas Schwung}
\ead{schwung.andreas@fh-swf.de}

\cortext[cor1]{Corresponding author.}
\address[label1]{Department of Automation Technology and Learning Systems, South Westphalia University of Applied Sciences, Lübecker Ring 2, Soest, 59494, Germany}
\address[label2]{Department of Artificial Intelligence and Data Science in Automation Technology, Hochschule D{\"u}sseldorf University of Applied Sciences, M{\"u}nsterstra{\ss}e 156, D{\"u}sseldorf, 40476, Germany}

\begin{abstract}
This paper presents a novel online transfer learning approach in state-based potential games (TL-SbPGs) for distributed self-optimization in manufacturing systems. The approach targets practical industrial scenarios where knowledge sharing among similar players enhances learning in large-scale and decentralized environments. TL-SbPGs enable players to reuse learned policies from others, which improves learning outcomes and accelerates convergence. To accomplish this goal, we develop transfer learning concepts and similarity criteria for players, which offer two distinct settings: (a) predefined similarities between players and (b) dynamically inferred similarities between players during training. The applicability of the SbPG framework to transfer learning is formally established. Furthermore, we present a method to optimize the timing and weighting of knowledge transfer. Experimental results from a laboratory-scale testbed show that TL-SbPGs improve production efficiency and reduce power consumption compared to vanilla SbPGs.
\end{abstract}

\begin{keyword}
Transfer learning, multi-objective optimization, game theory, smart manufacturing, distributed learning, machine learning
\end{keyword}

\end{frontmatter}

\section{Introduction}\label{sec:intro}

Modern manufacturing systems~\citep{Jan2023} are increasingly expected to be intelligent, agile, highly flexible, and adaptable, given the rapid and continuous changes in their production processes. To address these demands, game theory (GT) applied in dynamic games for engineering applications~\citep{Bauso2016} emerges as a promising approach. GT's collaborative characteristics are particularly suitable for self-learning in multi-agent systems (MAS), which require fast adaptation and reconfiguration while each agent's decision-making affects each other. Within industrial settings, GT-based approaches have demonstrated effectiveness in managing decision-making tasks within MAS, where multiple agents collaborate in a shared environment to tackle multi-objective optimization challenges~\citep{Schwung2020}. Despite its potential advantages, the practical implementation of GT in real manufacturing settings remains relatively limited, primarily due to the vast amounts of required training data and the lengthy duration of training~\citep{Schwung2020}. Particularly, learning from scratch is often impractical without access to highly precise digital system representations. Furthermore, the large-scale nature of most production systems worsens the time required for training~\citep{Wang2018, Wuest2016}. However, there are similarities within modules of the system, such as common objectives, hardware actuation, and control parameters, which can be utilized to reduce training time.

\begin{figure}[t]
 \centering
 \includegraphics[width=0.9\columnwidth,keepaspectratio]{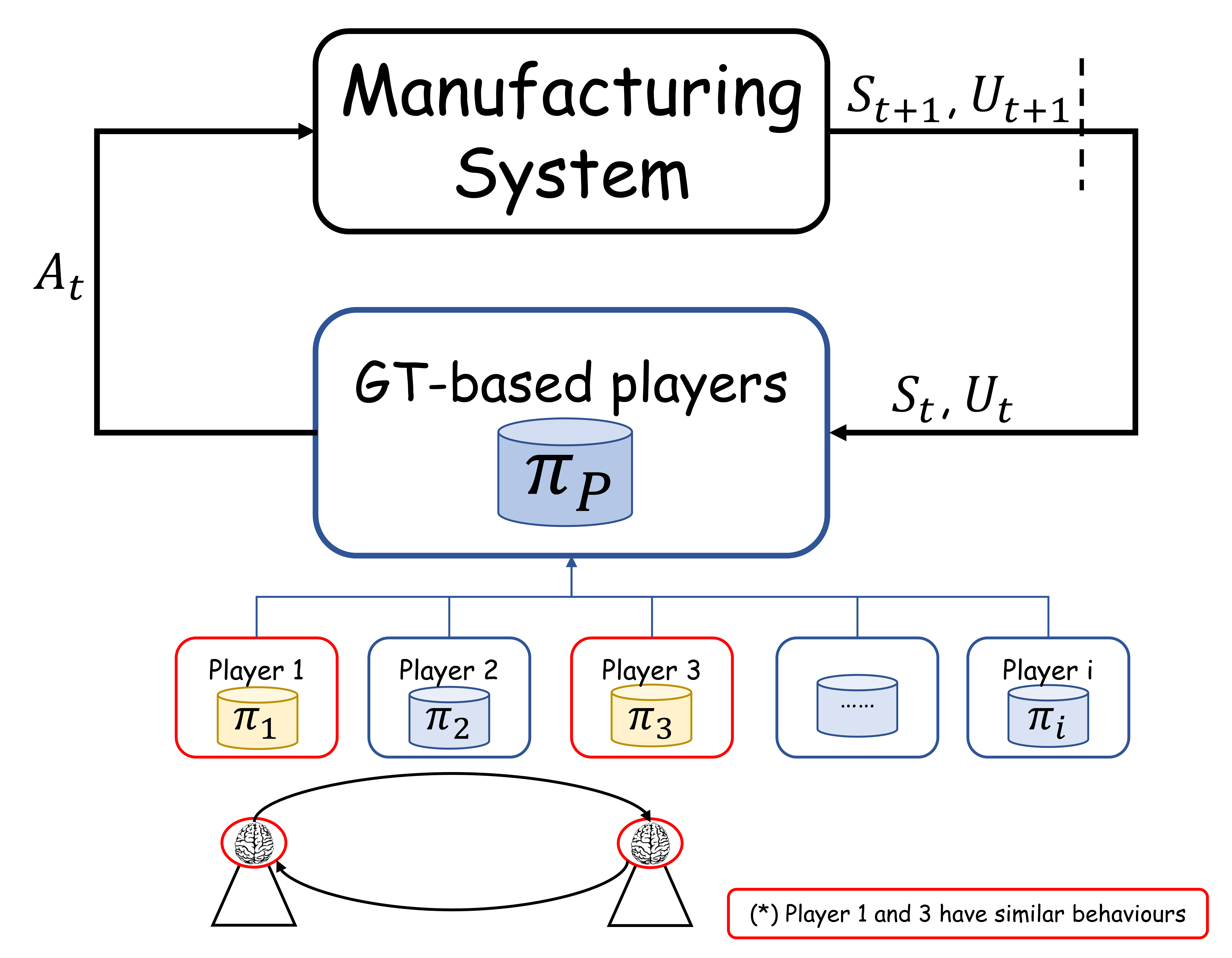}
\caption{Curriculum for transfer learning in GT-based learning.}
\label{fig:tl_in_gt}
\end{figure}

Therefore, in this paper, we propose the first approach that applies online transfer learning within state-based potential games (SbPGs) for distributed manufacturing systems, combining established GT learning with a similarity-based knowledge transfer mechanism across system modules. Our objective is to accelerate training duration and improve production efficiency beyond developed GT-based learning methods. Unlike previous GT-based learning methods, our transfer learning method employs communication signals as in~\citep{Yuwono2022} to facilitate knowledge transfer among players with common interests during training. Our approach aims to preserve the knowledge gained from solving specific cases and efficiently reuse it in different scenarios, thus saving resources. We claim that transfer learning accelerates player learning by extensively learning from the experiences of other players, as illustrated in Fig.~\ref{fig:tl_in_gt}.

Research on the implementation of transfer learning in industrial control and GT domains is limited. To our knowledge, transfer learning has not yet been applied in GT, including potential games. However, successful practical applications of transfer learning techniques in dynamic programming offer an understanding of how GT can be enhanced. Examples include unsupervised and supervised deep transfer learning for fault diagnosis~\citep{Zhao2021, Su2024}, transfer learning on evolutionary algorithms~\citep{Tan2021}, and transfer learning in reinforcement learning domains~\citep{Zhu2023}. In these cases, transfer setups typically rely on human decisions considering task similarities and are not straightforwardly executable. Moreover, most applications primarily focus on single-agent scenarios. In contrast, our approach targets multi-agent systems while maintaining a distributed player setup. Furthermore, our transfer learning occurs between training players in a parallel manner, rather than involving external or pre-trained agents. In addition, there is currently a lack of large-scale studies in manufacturing automation about transfer learning in self-optimizing production control systems, which would provide valuable real-world examples.

Our proposed approach contains two primary scenarios. Firstly, we propose a transfer learning framework with the assumption that prior analysis has identified similarities in players' behaviors. By establishing these similarities at the outset of the learning process, knowledge transfer among players can be pre-determined. Secondly, we consider a scenario where initial information about the similarities between players' behaviors is unavailable. To address this, we develop a methodology through a radial basis function network for measuring similarities among players and facilitating knowledge transfer between players in the absence of such information. The primary benefit of our approach is its ability to accelerate the learning of players' behaviors and performance. This is achieved through knowledge sharing among players, which enables faster convergence towards optimal solutions.

We provide formal proofs demonstrating that our proposed transfer learning frameworks fulfil the requirements of potential game-based learning, which ensures corresponding convergence guarantees~\citep{Zazo2016, Marden2009}. During our study, we observed that the timing and weighting of the transfer learning procedure between players during training are crucial factors. To address this challenge, we introduce a method that effectively manages the transfer learning process. Subsequently, we evaluate the performance of our transfer learning approaches using a laboratory testbed simulating a bulk good system, particularly in optimizing power consumption and throughput times. The experimental results demonstrate promising outcomes for improving production efficiency.

The main contributions of the paper are summarized as follows:
\begin{itemize}
\item We introduce TL-SbPGs, a novel transfer learning methodology for distributed optimization in SbPGs, which enables players to exchange and reuse experiences to optimize their policies.
\item We present two transfer learning approaches based on predefined player similarities: (1) the momentum-based (MOM) approach and (2) the sliding window (SW) approach. We provide formal proof that both approaches satisfy the requirements of SbPGs.
\item We propose an approach to handle unknown player similarities, using radial basis function networks. This method can also be applied independently for similarity measurement.
\item We introduce a method to optimize the timing and weighting of transfer learning, which demonstrates that these factors significantly affect the effectiveness of TL-SbPGs.
\item We validate the proposed methodologies through experiments on a laboratory testbed and larger-scale production systems, which show that TL-SbPGs with and without similarity measures deliver highly promising results across various production scenarios.
\end{itemize}

The paper is structured as follows: Section~\ref{sec:relwork} describes the related work of our topic. Section~\ref{sec:prob} provides the problem statement of the underlying distributed optimization problem for modular production units. Section~\ref{sec:GT} introduces the basic concepts of SbPGs. In Section~\ref{sec:meta}, we present a comprehensive explanation of various transfer learning approaches in TL-SbPGs. Then, Section~\ref{sec:results} provides results and comparisons of various transfer learning approaches in TL-SbPGs. Section~\ref{sec:conclusion} concludes the paper and delivers the future scope of the approach.

\section{Related work}\label{sec:relwork}

This chapter presents a discussion about related works, including self-learning in production systems, transfer learning for automation in manufacturing, and potential games.

\subsection{Self-learning in production systems}

Automation technology has gained significant attention in modern manufacturing systems, especially in the era of digitization, where automated production optimization plays a vital role~\citep{Jan2023}. Self-learning production systems are a prime example of this application, where systems can autonomously learn from their experiences, make independent decisions, and self-diagnose their conditions~\citep{Sahoo2022, Wang2021}. Various methods and applications have been established in this domain, which showcase the effectiveness of automation technology. For instance, Bayesian optimization has been successfully employed for auto-tuning throttle valves~\citep{Brosig2020}, knowledge-assisted reinforcement learning has proven beneficial for turbine control in wind farms~\citep{Zhao2020}, automated Programmable Logic Controller (PLC) code generation via evolutionary algorithms~\citep{Loppenberg2023}, vision-based reinforcement learning for bucket elevator process optimization~\citep{Chavan2024}, and further applications in production systems~\citep{Wang2021, Panzer2022}.

Recent research~\citep{Schwung2022, Yuwono2023a, Yuwono2025} has shown that GT in dynamic games outperforms multi-agent reinforcement learning in self-learning for distributed manufacturing systems, due to its cooperative decision-making strategies. For example, in~\citep{Lowe2017, Foerster2018}, they introduce deep multi-agent reinforcement learning with a centralized critic and a central training, decentralized deployment paradigm. While this approach requires no communication during operation, it assumes full access to the state space during training, which is an unrealistic condition in real-world applications. Actor-critic frameworks, as in~\citep{Schwung2021}, also fall short compared to GT-based learning due to limited coordination among agents and the absence of guaranteed convergence to a Nash equilibrium. In addition, building upon the previous research, where communication signals between players in GT are enabled~\citep{Yuwono2022}, they have yet to fully explore the potential benefits of such communication among actuators in production systems. In this study, we aim to utilize this capability to accelerate the learning process while enhancing performance. Our prior studies provide a robust foundation for the current research. Despite the promise of self-learning systems, effectively reusing prior knowledge remains challenging, which motivates our exploration of transfer learning in the following subsection.

\subsection{Transfer learning for automation in manufacturing}
Transfer learning is an advanced technique in the machine learning area and has emerged as a promising paradigm for enhancing automation in manufacturing systems by leveraging knowledge from related tasks or domains~\citep{Niu2020, Neyshabur2020}. Transfer learning facilitates the application of knowledge acquired from source tasks to accelerate the learning process of new tasks~\citep{Zhu2023, Silver2016}. This approach has proven effective in domains like medical imaging and cybersecurity, where transfer learning and explainable artificial intelligence support real-world decision systems through feature reuse, domain adaptation, training efficiency, and interpretability~\citep{Sakirin2025, Salloum2025}. In practical scenarios, the concept of transfer learning extends beyond a singular activity, which enables solutions to diverse cases such as cross-domains, cross-environments, or cross-phases~\citep{Cruz2016}. In recent literature, researchers have explored various transfer learning techniques to address the challenges of adapting models trained on source domains to target domains with limited or no labelled data~\citep{Hosna2022}. For instance, in the context of robotics and automation, transfer learning has been applied to tasks such as object recognition~\citep{Murali2022}, manipulation and scheduling~\citep{Tobin2017, Andrychowicz2020}, as well as motion planning~\citep{Wen2021}, where pre-trained models are fine-tuned or adapted to specific manufacturing environments.

Another transfer learning approach involves teacher-student learning, where traditional operations research solvers act as teachers to guide reinforcement learning algorithms in solving robot scheduling problems~\citep{Loeppenberg2024}. Moreover, transfer learning has also been utilized to transfer knowledge between different manufacturing processes or production lines, which enables more efficient utilization of resources and faster deployment of automation solutions. However, there are no large-scale studies about transfer learning in self-optimizing production control systems that can represent a real-world example, as demonstrated in our study. Additionally, the majority of applications primarily focus on single-agent and offline scenarios. In contrast, meta-learning aims to rapid adaptation to new environments by generalizing across tasks~\citep{Chen2021}, but it requires extensive meta-training and assumes consistent task distributions~\citep{Vettoruzzo2024}, which limits its use in dynamic and heterogeneous manufacturing settings. Our GT-based approach avoids these assumptions and adapts via inter-agent coordination without prior task knowledge.

\subsection{Potential games}
Potential Game (PG) was originally introduced in economics~\citep{Monderer1996}, but nowadays has been increasingly applied in engineering tasks~\citep{Bauso2016}. For instance, distributed parameter estimation of sensors in a non-convex problem~\citep{Ampeliotis2019} and maximizing system throughput of transmission rate~\citep{Kato2023} have been reported. As in~\citep{Monderer1996}, the potential function $\phi$ aids in achieving local or global optimality of the learned system and facilitates convergence studies. Additionally, incorporating various learning algorithms can ensure the convergence of a system to a Nash equilibrium (NE). However, the major challenge in PG-based optimization for a self-learning system is establishing a strategy that guarantees an untroubled alignment between the potential function and the local objective functions (utilities $U_i$) for each player $i$. Hence, to incorporate state information in dynamic PGs (DPGs)~\citep{Zazo2016}, SbPGs have been developed in~\citep{Marden2012} for the first time. Then, SbPGs have been implemented in distributed production systems as an algorithm for self-optimizing the systems~\citep{Schwung2020}. However, the fact is that the learning of the distributed policies is mostly conducted from scratch, resulting in lengthy training periods and is barely applicable in the real environment. In~\citep{Yuwono2022}, they extensively discussed communication and memory within SbPGs, yet the exploration of the full potential of communication remains unexplored. Additionally, an algorithm that integrates online model-based learning within SbPGs has been developed in~\citep{Yuwono2023a, Yuwono2023b}. Furthermore, in previous related work~\citep{Schwung2022}, transfer learning between PLC code and GT players has been explored using a teacher-student learning framework, where the PLC code acted as a teacher guiding GT players to mimic its behaviour. Existing transfer learning techniques typically assume the presence of a pre-trained agent, whose knowledge is transferred to other agents after its training is completed. In contrast, this study focuses on transfer learning exclusively between GT players, which implements an online approach that enables information exchange during the learning process and incorporates similarity measures.

Given the advantages and challenges of PGs and SbPGs in distributed optimization, we now define the specific problem addressed in this work, centered around modular and decentralized manufacturing systems.

\section{Problem statement}\label{sec:prob}

This section offers a detailed problem formulation embedded in the context of the fully distributed manufacturing system under consideration. Advanced manufacturing systems demand complex functionality to facilitate fast and highly adaptable production processes. Such capabilities are realized through a modular system structure comprising both hardware and software architectures, which enables inclusion, exclusion, or interchangeability of each module as per necessity. In a MAS setting, the development of distributed control systems must occur at the software level~\citep{Leitao2016}.

\begin{figure}[t]
 \centering
 \includegraphics[width=1.00\columnwidth,keepaspectratio]{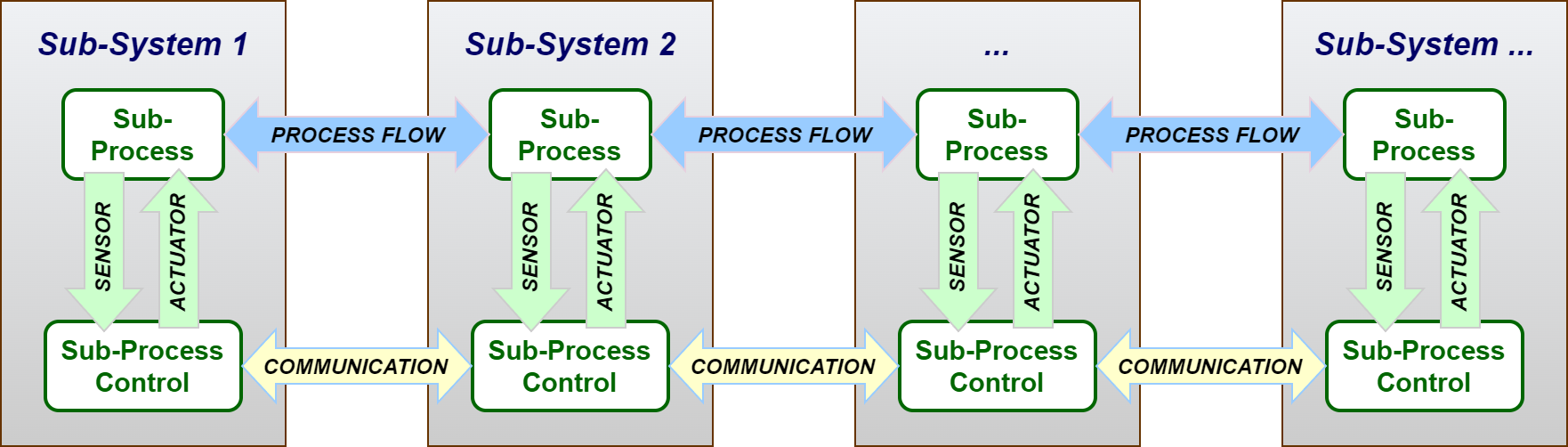}
\caption{Considered a distributed system structure including several subsystems with their own control and communication systems.}
\label{fig:sysstruct}
\end{figure}

Therefore, this study focuses on distributed, modular systems that are partitioned into several subsystems, each equipped with its local control system, as illustrated in Fig.~\ref{fig:sysstruct}. Our objective is to optimize the system in a fully distributed manner, without dependency on a centralized instance. This approach allows for flexible, scalable, and generally reusable execution across various modules through instantiations.

\begin{figure}[t]
 \centering
 \includegraphics[width=1.00\columnwidth,keepaspectratio]{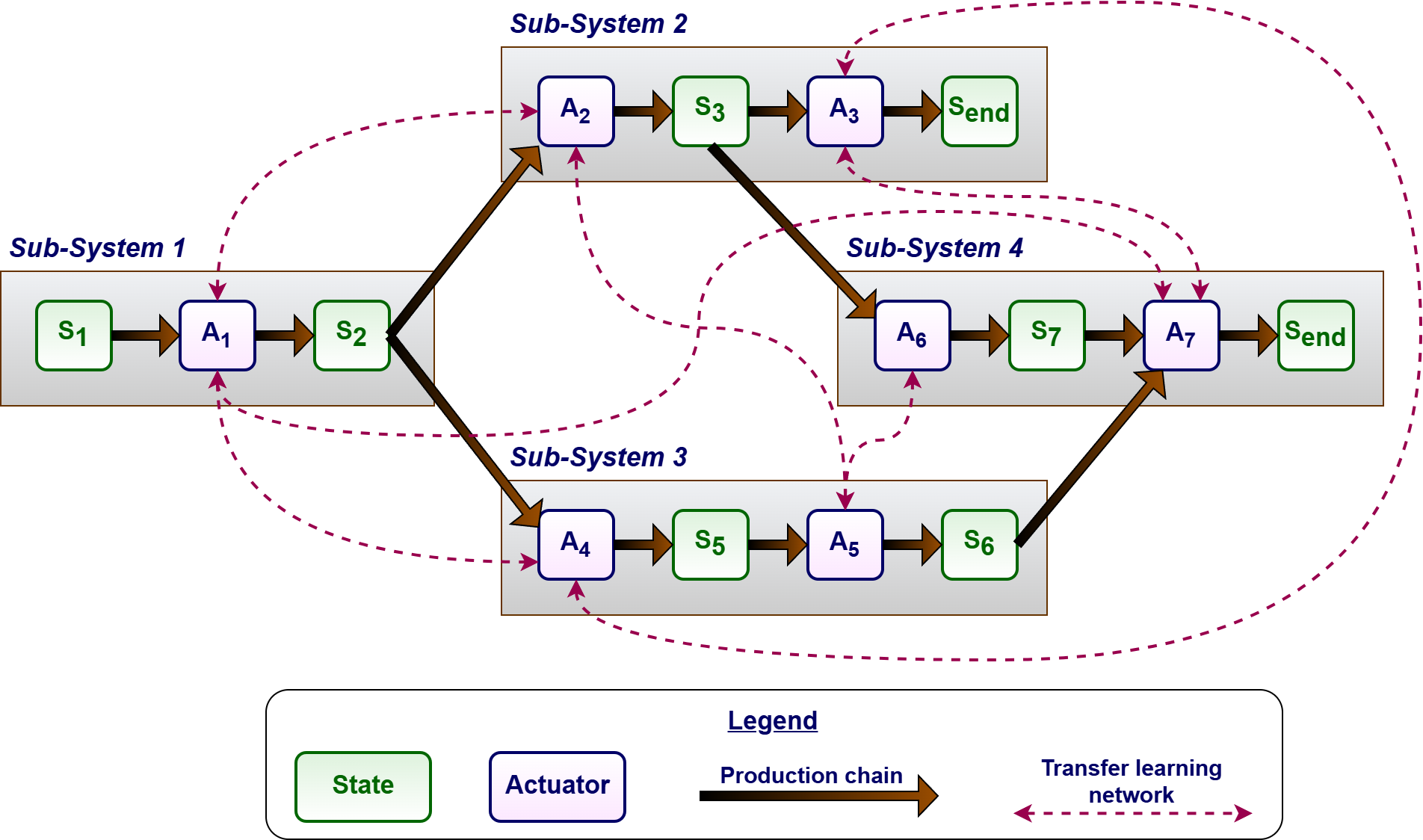}
\caption{Production chain schematic with transfer learning networks on serial and parallel connected sub-systems.}
\label{fig:prodchain}
\end{figure}

Each subsystem interacts with its control system by exchanging local signals. Additionally, there are two connections linking different subsystems, which are one at the communication level and another at the process level. The communication level is facilitated by communication interfaces, such as Ethernet, fieldbus, or wireless connections. These communication interfaces enable modules to be uniquely identified within the production environment, which facilitates knowledge transfer via communication signals in our study. We assume that the transfer learning process occurs in both serial and serial-parallel process chains, as depicted in Fig.~\ref{fig:prodchain}. Similar to~\citep{Schwung2020}, we consider that the production chain is represented by an alternating arrangement of actuators that operate on the process (e.g., motors, pumps, conveyors, pumps) and physical states representing the process status. These actuators are expected to exhibit both discrete and continuous operational behaviors, which form a hybrid actuation system. 

We represent and analyze the considered distributed system via graph theory~\citep{Yamamoto2015}, as further explained in~\citep{Schwung2020}. We define a set of actuators $\mathcal{N}={1;\ldots;N}$ and a set of states $\mathcal{S}\subset \mathbb{R}^m$ with a corresponding set of actions $\mathcal{A}_i \subset \mathbb{R}^c \times \mathbb{N}^d$ with continuous and/or discrete behaviour. We describe the production chain in the form of a directed graph with $|\mathcal{S}|$ states and $N$ actuator nodes. We assume a dynamic series of states and actuators, such that the set of edges $\mathcal{E}$ omits edges $e=(s_i,s_j)$ and $e=(A_i,A_j)$ with $s_i, s_j \in \mathcal{S}$ and $A_i, A_j \in \mathcal{N}$. Moreover, two sets of surrounding states are determined for every actuator, including the group of subsequent states $\mathcal{S}^{A_i}_{next}=\{s_j \in \mathcal{S}|\exists e=(A_i,s_j)\in \mathcal{E}\}$ and prior states $\mathcal{S}^{A_i}_{prior}=\{s_j \in \mathcal{S}|\exists e=(s_j,A_i)\in \mathcal{E} \}$. The functions of the local utility is characterized as $U_i=f(S^{A_i}, a_i)$, in which $a_i \in \mathcal{A}_i$ and ${S}^{A_i} \in \mathcal{S}^{A_i} = \mathcal{S}^{A_i}_{prior} \cup \mathcal{S}^{A_i}_{next} \cup \mathcal{S}^{g}$. The group of states $\mathcal{S}^{g}$ is utilized to determine global objectives.

In our study, we enable communication interfaces to facilitate transfer learning between players via communication signals, which leads to the introduction of an auxiliary utility function $H_{i,j}$ that encourages transfer between player $i$ and player $j$. We intend to subsequently integrate this into the local utility function of each player, a topic to be addressed in Section~\ref{sec:meta}. However, in this setup, the multiple utility functions may share states, and the overarching objective is to maximize overall utility, as outlined below:
\begin{align}\label{eq:optproblem}
\max_{{a}\in \mathcal{A}} \phi({a},{S}) = \sum_{i=0}^{N} U_i(a_i, S^{A_i})
\end{align}
by jointly accelerating local utilities $U_i(a_i, S^{A_i})$ of the system~\citep{Zazo2016}, where $\phi$ denotes a global objective function, known as the potential function. As a general framework for this setup, we utilize SbPGs for distribution optimization, as explained in the next section.

This generalized production scenario is applicable across industries such as pharmaceuticals, food production, and chemical plants, as well as in energy systems like smart grids and logistics, provided the problem setup involves distributed optimization, as explained in~\citep{Schwung2020b}. Decentralized production lines, which require flexible and scalable solutions, are increasingly common in modern industrial environments. To be noted, the assumption that the production chain comprises only sequences of states and actions is not restrictive. A sequence of more than two states can be reformulated into a common state vector, and similar arguments apply to actions. In the manufacturing context, local utilities can represent multi-strategic goals such as minimizing throughput time, idle time, storage requirements, setup and changeover time of machines, reducing production rejects and rework, and optimizing power consumption. Practical applications often involve a combination of different multi-objectives, which leads to multi-objective optimization problems.

To effectively model the distributed control problem formulated above, we apply the SbPGs, whose key concepts and formal properties are detailed in the following section.

\section{Basics of state-based potential games}\label{sec:GT}

In this section, we introduce SbPGs, which were mentioned in the previous section, to formulate distribution optimization within modular manufacturing units. SbPGs are the inheritance of PGs~\citep{Monderer1996}, which refer to a type of strategic game that relies on the potential function $\phi$ and additional state information.

In PGs, we introduce a set of $N$ players with a strategy set for each player \textit{i} organized as discrete actions $a_i \in \mathbb{N}^{d_i}$ and/or continuous actions $a_i \in \mathbb{R}^{d_i}$, and a potential function $\phi$ on joint actions $a=a_1 \times \ldots \times a_N$ yields the global optimization problem $\max_{{a} \in  \mathcal{A}} \phi(a)$ that split among the players~\citep{Monderer1996}:  
\begin{definition}
	A strategic-form game $\Gamma(\mathcal{N}, \mathcal{A}, \{U_i\}, \phi)$ establishes an (exact) PG if a global function $\phi:  a \rightarrow \mathcal{R}^{d_i}$ can be found conforming to the condition
	\begin{align}\label{eq:utilitydef}
	U_i(a_i, {a}_{-i}) - U_i({a}^{\prime}_i, {a}_{-i}) = \phi(a_i, {a}_{-i}) - \phi({a}^{\prime}_i, {a}_{-i}),
	\end{align} 
	for any $i=1, \ldots, N$, $a_i \in \mathcal{A}_i$ and $ {a}_{-i} \in \mathcal{A}_1 \times  \ldots \times \mathcal{A}_{i-1} \times \mathcal{A}_{i+1} \times \ldots \times \mathcal{A}_N$.   
\end{definition}    
To reach the convergence to an NE, a potential function maximizer from the set $a^* = \text{argmax}_{ {a} \in  \mathcal{A}} \phi(a)$ is set up.

\begin{definition}
    If the action $a_i^*$ for each player $i \in \mathcal{N}$ is the best response to the joint action of other players $a_{-i}^*$, then the joint action $a^* \in \mathcal{A}$ is an NE, specifically
	\begin{align}\label{eq:nashequi}
	U_i(a_i^*, {a}_{-i}^*) = \max_{a_i \in A_i} U_i(a_i, {a}_{-i}^*).
	\end{align}
\end{definition}
Hence, no player can benefit by unilaterally deviating from the NE action profile. However, the limitation of PGs in the engineering area is the potential to incorporate state information about the environment with the games themselves, which is solved by the idea of SbPGs~\citep{Marden2012}.

SbPG~\citep{Marden2012} has an underlying state space that allows the environment's dynamics to be considered during the game, hence an instance of DPGs~\citep{Zazo2016}. The utility function of each player is both action- and state-dependent, and the incorporation of state information varies depending on the setting, i.e., it ranges from action memory~\citep{Marden2012} for supporting experience learning to the state information from the environment into the PG. The formal definition of an SbPG is as follows:
\begin{definition}
	A game $\Gamma(\mathcal{N}, \mathcal{A}, \{U_i\}, {S}, P, \phi)$ constitutes an SbPG if a global function $\phi:  a \times  {S} \rightarrow \mathcal{R}^{d_i}$ can be found that for every state-action-pair $[s,a]\in {S} \times \mathcal{A}$ conforms to the conditions
	\begin{align}\label{eq:potcondsbpg}
	U_i(s,a_i) - U_i(s,{a}^{\prime}_i, {a}_{-i}) = \phi(s,a_i) - \phi(s,{a}^{\prime}_i, {a}_{-i})
	\end{align}
	and
	\begin{align}\label{eq:condsbpg}
	\phi(s^{\prime},a_i) \geq \phi(s,a_i)
	\end{align}
	for any state $s^{\prime}$ in the state transition process $P(s,a)$.
\end{definition} 

The most crucial characteristic of (exact) PGs, which is the convergence to equilibrium points, remains operative for SbPGs~\citep{Marden2012}. Different conditions have been derived to prove the existence of an SbPG for a given game setting~\citep{Zazo2016}. The following condition must be satisfied to verify the existence of a PG and to determine the utility function:
\begin{lemma}\label{lemma1}~\citep{Zazo2016}
	A game $\Gamma(\mathcal{N}, \mathcal{A}, \{U_i\}, {S}, P, \phi)$ is a DPG if the players' utilities satisfy the following conditions:
	\begin{align}\label{eq:dpg1}
	\frac{{\partial}^2U_i(s_i,a)}{{\partial}a_j{\partial}s_m}=\frac{{\partial}^2U_j(s_j,a)}{{\partial}a_i{\partial}s_n},
	\end{align}
	\begin{align}\label{eq:dpg2}
	\frac{{\partial}^2U_i(s_i,a)}{{\partial}s_n{\partial}s_m}=\frac{{\partial}^2U_j(s_j,a)}{{\partial}s_m{\partial}s_n},
	\end{align}
	\begin{align}\label{eq:dpg3}
	\frac{{\partial}^2U_i(s_i,a)}{{\partial}a_j{\partial}a_i}=\frac{{\partial}^2U_j(s_j,a)}{{\partial}a_i{\partial}a_j},
	\end{align}
	$\forall i,j \in \mathcal{N}$, $\forall m \in S^{A_i}$ and $\forall n \in S^{A_j}$.\\
\end{lemma}


Furthermore, a convex combination of two utility functions constitutes an SbPG if each individual utility function constitutes an SbPG, as proved in
\begin{lemma}\label{lemma2}~\citep{Schwung2022}
    A game $\Gamma(\mathcal{N}, \mathcal{A},\{\beta_1 U^{(1)}_i+\beta_2 U^{(2)}_i\}, {S}, P, \beta_1 \phi_1+\beta_2 \phi_2)$ with $\beta_1, \beta_2>0$ and $\beta_1 + \beta_2 = 1$ is an SbPG, if $\Gamma_1(\mathcal{N}, \mathcal{A},\{U^{(1)}_i\}, {S}, P, \phi_1)$, $\Gamma_2(\mathcal{N}, \mathcal{A},\{U^{(2)}_i\}, {S}, P, \phi_2)$ are SbPGs.
\end{lemma}

Lemma~\ref{lemma2} is formally for convex combinations, but it can be easily extended to the case where $\beta_1, \beta_2$ are arbitrary real numbers.

With the SbPG structure in place, we are now positioned to extend it with transfer learning capabilities. The next section introduces TL-SbPGs, our proposed approach to accelerate and improve learning via knowledge reuse and transfer among players.

\section{Transfer learning of state-based potential games}\label{sec:meta}

In this section, we present a novel method called TL-SbPGs, which combines transfer learning approaches with SbPGs. TL-SbPGs facilitate knowledge transfer between players with similar behaviors during training to accelerate the learning process by also learning extensively from experiences gained from other players, rather than discarding them, and leading to faster convergence towards optimal solutions.

\begin{figure}[t]
 \centering
 \includegraphics[width=1.00\columnwidth,keepaspectratio]{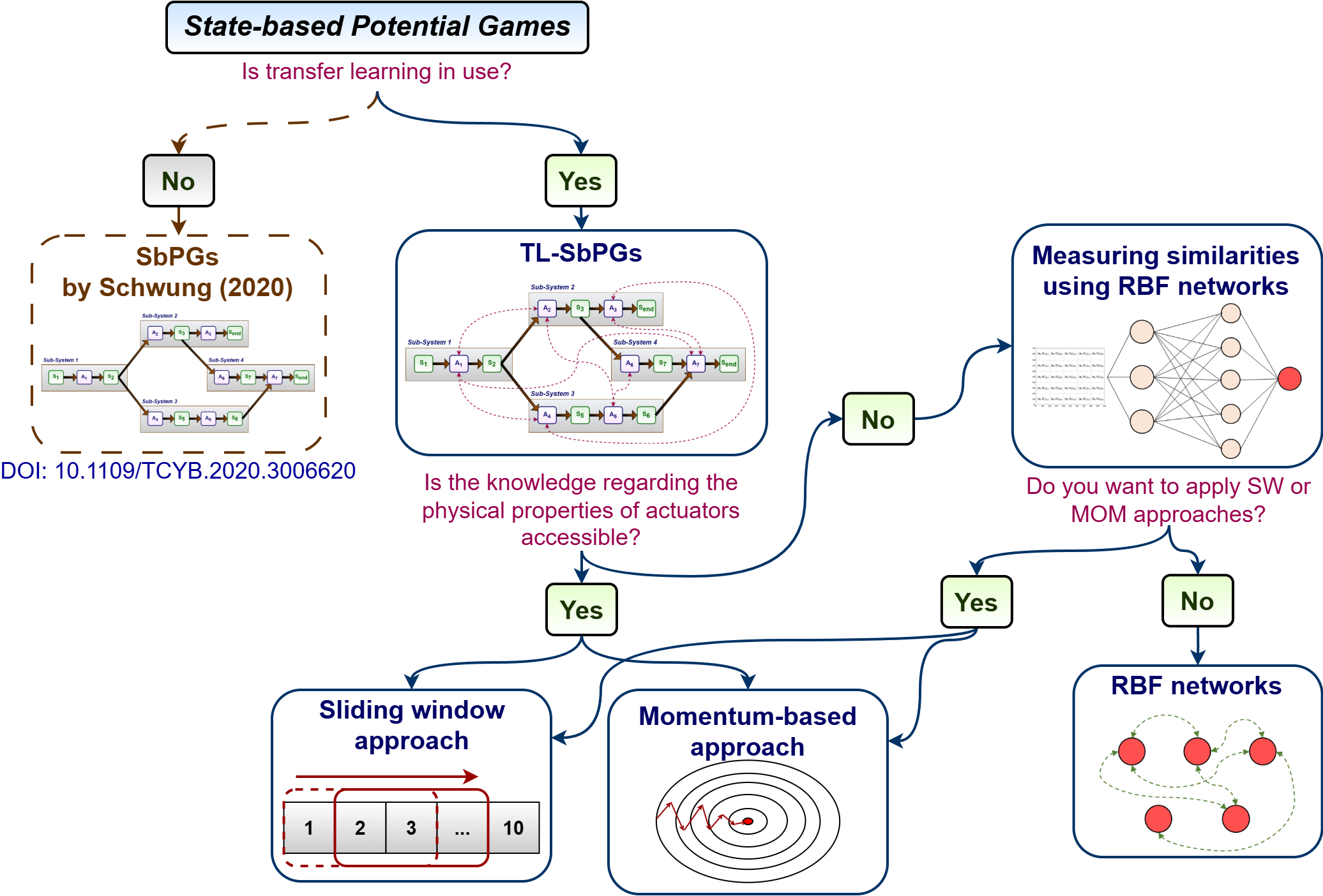}
\caption{An overview of TL-SbPGs for distributed optimizations.}
\label{fig:tl_sbpg_overview}
\end{figure}

Fig.~\ref{fig:tl_sbpg_overview} describes an overview of the proposed TL-SbPGs for distributed optimizations. We explore two scenarios in which TL-SbPGs can be implemented: (1) when the similarity between players is given beforehand, e.g., when physically similar actors for physically similar functions, and (2) when such similarities must be inferred during training. For the latter scenario, we derive suitable similarity measures between the actuators, thereby respecting the communication constraints required in real-life applications. Additionally, we provide convergence guarantees for the learning process by proving that the proposed method satisfies the requirements of SbPGs while incorporating additional objectives.

The section is divided into three subsections, which are (A) the transfer learning frameworks for the first scenario, which include SW and MOM approaches, (B) the transfer learning frameworks for the second scenario, in which similarity measures must be inferred during training through radial basis function networks, and (C) the update mechanism for the TL-SbPGs.

\subsection{TL-SbPGs with pre-defined similarities}\label{sec:tlf}

We start with the assumption that knowledge regarding the physical properties of the process is available, which allows for the pre-definition of similarities between actuators and their respective functions within the process. Additionally, without loss of generalization, we assume that all variables, i.e., states and actions are normalized to the interval $[0,1]$. Further, we consider a suitable discretization of the state-action space as in~\citep{Schwung2020}.

We start by introducing a method for measuring similarities that facilitates effective transfer learning between players. To achieve this, we introduce an auxiliary utility function that is designed to encourage the transfer of knowledge. To this end, we propose to modify the utility function of SbPGs to allow the transfer learning between actuators, as follows:
\begin{align}\label{eq:newgame}
\tilde{U}_i(a_i, S^{A_i}) = U_i(a_i, S^{A_i}) \!-\! \alpha_{\text{TF}} H_{i,j}(a_i, a_j)
\end{align}
where $U_i(a_i, S^{A_i})$ is the original utility function of player $i$ as defined in Section~\ref{sec:GT}, $\alpha_{\text{TF}}$ is a weight parameter of the transfer learning and $H_{i,j}(a_i, a_j)$ denotes a suitably defined auxiliary utility function to encourage transfer between player $i$ and $j$.

Eq.~\eqref{eq:newgame} requires the definition of a suitable, potentially adaptive weight parameter $\alpha_{\text{TF}}$ as well as the auxiliary utility function $H_{i,j}(a_i, a_j)$, for which we introduce two different variants. Specifically, we propose an SW approach as well as a MOM approach to measure the similarity of the two action maps. We further formally prove the existence of an SbPG for both settings in the following.

\subsubsection{Sliding window approach}\label{sec:sw_app}

The first variant is based on an SW approach and yields, as follows:
\begin{align}\label{eq:swloss}
H^{\text{SW}}_{i,j}(a_i, a_j) = \sum_{k=0}^{H-1} (a_{i,t-k} - a_{j,t-k})^2,
\end{align} 
where $H$ is the horizon of the SW. The determination of the horizon length is contingent upon the learned systems, and it can also be scientifically defined through methods such as hyperparameter tuning, for instance, via grid search~\citep{Bergstra2013}. Thus, the loss is accumulated not only based on the actual time frame but also on the horizon. 

In this subsection, we aim to formally prove that the modified utility function in Eq.~\eqref{eq:newgame} constitutes the SbPG setting. As such, the equation consists of two parts, i.e., the original utility function and the auxiliary loss function. Lemma~\ref{lemma3} formally proves the existence of the SbPGs for the SW approach:
\begin{lemma}\label{lemma3}  
	Given a game $\Gamma(\mathcal{N}, \mathcal{A}, \{H^{\text{SW}}_{i,j}\}, {S}, P, \phi)$ with \textit{$H^{\text{SW}}_{i,j}$} shown in Eq.~\eqref{eq:swloss} constitutes the SbPGs.
\end{lemma}

\begin{proof} For the original utility function without additional cost criteria~\citep{Schwung2020} has been proven that it constitutes the SbPGs setting under reasonable assumptions. Furthermore, according to Lemma~\ref{lemma2}, the convex combination of two SbPG-capable utilities results in SbPG-capable utilities. Hence, we are left to prove that the auxiliary utility function $H_{i,j}(a_i, a_j)$ constitutes an SbPG. As the policy is particularly dependent on the actions of two players \textit{$a_i, a_j$} and does not involve any states \textit{$S^{A_i}$}, we can conclude that conditions~\eqref{eq:dpg1} and~\eqref{eq:dpg2} are fulfilled. For condition~\eqref{eq:dpg3}, we prove it, as follows:
\begin{align}\label{eq:lemma3_1}
\begin{split}
    &\frac{{\partial}^2H^{\text{SW}}_{i,j}(a_i, a_j)}{{\partial}a_i{\partial}a_j} 
    = \frac{\sum_{k=0}^{H-1} {\partial}^2(a_{i,t-k} - a_{j,t-k})^2}{{\partial}a_i{\partial}a_j} \\
    & = \frac{\sum_{k=0}^{H-1} (-1)^2{\partial}^2(a_{j,t-k} - a_{i,t-k})^2}{{\partial}a_i{\partial}a_j} \\
    & =\frac{\sum_{k=0}^{H-1} {\partial}^2(a_{j,t-k} - a_{i,t-k})^2}{{\partial}a_j{\partial}a_i}
    = \frac{{\partial}^2H^{\text{SW}}_{j,i}(a_j, a_i)}{{\partial}a_j{\partial}a_i},
\end{split}
\end{align}
$\forall i,j \in \mathcal{N}$.
Since the criteria for constituting an SbPG are fulfilled, we can summarize that the transfer learning objective can be incorporated into SbPGs.
\end{proof}

As an alternative to the SW approach, we present a MOM method that incorporates historical similarity trends more smoothly over time.

\subsubsection{Momentum-based approach}\label{sec:mom_app}

Alternatively, we define the auxiliary utility function based on the MOM method as follows:
\begin{equation}\label{eq:momloss}
H^{\text{MOM}}_{i,j,t}(a_i, a_j) = \alpha_{MOM} H^{\text{MOM}}_{i,j,t-1}(a_i, a_j) + (1-\alpha_{MOM}) (a_{i,t} - a_{j,t})^2,
\end{equation}
where for $t=0$, the auxiliary utility function is transformed, as follows:
\begin{align}\label{eq:momloss2}
H^{\text{MOM}}_{i,j,0}(a_i, a_j) = (a_{i,0} - a_{j,0})^2,
\end{align} 
where $\alpha_{MOM}$ represents a fixed weight that balances the optimal contributions of the previous and current losses. The weight parameter $\alpha_{MOM}$ determines the proportion of the previous auxiliary utility function $H^{\text{MOM}}_{i,j,t-1}(a_i, a_j)$ that is retained for the next step calculation, a feature that is not present in the SW approach. Lemma~\ref{lemma4} formally proves the existence of the SbPGs for the MOM approach:
\begin{lemma}\label{lemma4}  
	Given a game $\Gamma(\mathcal{N}, \mathcal{A}, \{H^{\text{MOM}}_{i,j,t}\}, {S}, P, \phi)$ with \textit{$H^{\text{MOM}}_{i,j,t}$} shown in Eq.~\eqref{eq:momloss} constitutes an SbPG.
\end{lemma}

\begin{proof} 
Similar to the SW approach, we resort to the proof of the existence of an SbPG for the auxiliary loss~\eqref{eq:momloss} due to Lemma~\ref{lemma2}. Again, condition~\eqref{eq:dpg1} and~\eqref{eq:dpg2} are trivially fulfilled. However, as the momentum-based loss function takes the previous losses into account, it forms an infinite arithmetic sequence. Therefore, we formally prove the infinite arithmetic sequence using mathematical induction. The proof consists of two steps, namely the base case and the inductive step.

The base case occurs at $t=0$, described by Eq.~\eqref{eq:momloss2}. For condition~\eqref{eq:dpg3}, we prove it, as follows:
\begin{align}
    \begin{split}
        & \frac{{\partial}^2H^{\text{MOM}}_{i,j,0}(a_i, a_j)}{{\partial}a_i{\partial}a_j} = \frac{{\partial}^2(a_{i,0} - a_{j,0})^2}{{\partial}a_i{\partial}a_j} \\
        & = \frac{{\partial}^2(-1)^2(a_{j,0} - a_{i,0})^2}{{\partial}a_i{\partial}a_j} \\ 
        & =\frac{{\partial}^2(a_{j,0} - a_{i,0})^2}{{\partial}a_j{\partial}a_i} =
        \frac{{\partial}^2H^{\text{MOM}}_{j,i,0}(a_j, a_i)}{{\partial}a_j{\partial}a_i},
    \end{split}
\end{align}
$\forall i,j \in \mathcal{N}$.

Next, for $t\neq0$, the momentum-based loss function in Eq.~\eqref{eq:momloss} consists of two objectives, such as the previous and actual losses. We have to prove condition~\eqref{eq:dpg3}, where \textit{$a_i, a_j \in [0,1]^2$}, as follows:
\begin{align}\label{eq:lemma4_1}
    \begin{split}
\frac{{\partial}^2H^{\text{MOM}}_{i,j,t}(a_i, a_j)}{{\partial}a_i{\partial}a_j}=\alpha_{MOM}\frac{{\partial}^2H^{\text{MOM}}_{i,j,t-1}(a_i, a_j)}{{\partial}a_i{\partial}a_j} + \frac{{\partial}^2(1-\alpha_{MOM}) (a_{i,t} - a_{j,t})^2}{{\partial}a_i{\partial}a_j} \\
\stackrel{!}{=}
\frac{{\partial}^2H^{\text{MOM}}_{j,i,t}(a_j, a_i)}{{\partial}a_j{\partial}a_i}=\alpha_{MOM}\frac{{\partial}^2H^{\text{MOM}}_{j,i,t-1}(a_j, a_i)}{{\partial}a_j{\partial}a_i} + \frac{{\partial}^2(1-\alpha_{MOM}) (a_{j,t} - a_{i,t})^2}{{\partial}a_j{\partial}a_i},
\end{split}
\end{align}
$\forall i,j \in \mathcal{N}$.
This can be proven in three steps. First, we have again a combination of two objectives; hence, Lemma~\ref{lemma2} applies. Second, according to the transfer step in mathematical induction, if $H^{\text{MOM}}_{i,j,t}$ is an SbPG, then $H^{\text{MOM}}_{i,j,t+1}$ is also an SbPG. As $H^{\text{MOM}}_{i,j,0}$ has been proven to be an SbPG, also $H^{\text{MOM}}_{i,j,t}$ remains a PG. Third, we prove that the second objective of the momentum-based loss function  constitutes an SbPG:
\begin{align}
    \begin{split}
&\frac{{\partial}^2(1\!-\!\alpha_{MOM}) (a_{i,t} \!-\! a_{j,t})^2}{{\partial}a_i{\partial}a_j} \!=\! \frac{{\partial}^2(1\!-\!\alpha_{MOM}) (\!-\!1)^2(a_{j,t} \!-\! a_{i,t})^2}{{\partial}a_j{\partial}a_i} \\
&=\! \frac{{\partial}^2(1\!-\!\alpha_{MOM}) (a_{j,t} \!-\! a_{i,t})^2}{{\partial}a_j{\partial}a_i} \!=\! -2(1\!-\!\alpha_{MOM}),
\end{split}
\end{align}
$\forall i,j \in \mathcal{N}$. Hence, we have:
\begin{align}\label{eq:lemma4_4}
\frac{{\partial}^2H^{\text{MOM}}_{i,j}(a_i, a_j)}{{\partial}a_i{\partial}a_j}=\frac{{\partial}^2H^{\text{MOM}}_{j,i}(a_j, a_i)}{{\partial}a_j{\partial}a_i},
\end{align}
$\forall i,j \in \mathcal{N}$, which concludes the proof.
\end{proof}

In summary, both transfer learning objectives can be integrated into SbPGs. We note that the above result has favourable implications for the communication requirements necessary to achieve convergence. Specifically, to ensure convergence, it suffices to establish an undirected interaction graph between players assumed to be similar. Consequently, each player must communicate with at least one other player, but is not restricted to only one player. We remark that the variants of cost criteria represent a form of consensus protocol. In previous research~\citep{Marden2009}, it has been demonstrated that certain forms of consensus protocols fulfil the PG setting, although not the SbPG setting.

To further refine the transfer process, we introduce a dynamic weighting mechanism that governs when and how strongly knowledge transfer should be applied.

\subsubsection{Adaptive weight parameter}\label{sec:adappara}

As an additional contribution in this paper, we address two crucial questions regarding transfer learning between players: (1) when the transfer learning process should start, and (2) how much relevant knowledge each player can share with others. To tackle these questions, we propose a method to dynamically determine the value of the weight parameter $\alpha_{\text{TF}}$ in Eq.~\eqref{eq:newgame}. This parameter is important in initiating and quantifying the significance of knowledge transfer. Our objective is to establish the timing and weighting factors for the transfer of knowledge between players, which ensures that the transfer learning process starts at an appropriate time to avoid premature initiation and the transmission of irrelevant information.

First, to determine when to initiate the transfer learning process, we introduce a transfer learning threshold $\beta_{\text{TF}}$ as a kick-off indicator. The transfer learning process is triggered once the exploration rate $\epsilon$ of the learning player falls below the threshold $\beta_{\text{TF}}$. Specifically, when $\epsilon > \beta_{\text{TF}}$, the transfer learning parameter $\alpha_{\text{TF}}$ is set to 0, which indicates that no transfer learning occurs. This approach ensures that transfer learning is not initiated during periods of high exploration, thereby avoiding disruption of the exploration phase with excessive knowledge transfer. 

\begin{figure}[t]
\centering
\raisebox{-\height}{\includegraphics[width=0.45\textwidth]{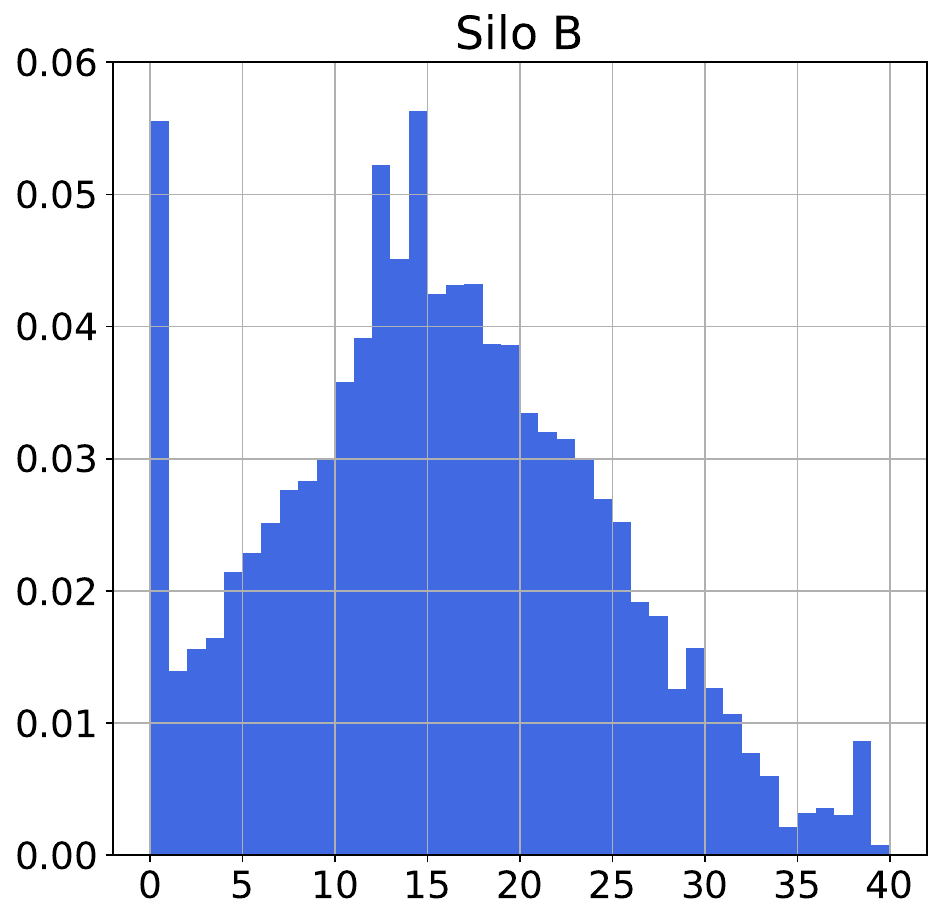}}
\raisebox{-\height}{\includegraphics[width=0.45\textwidth]{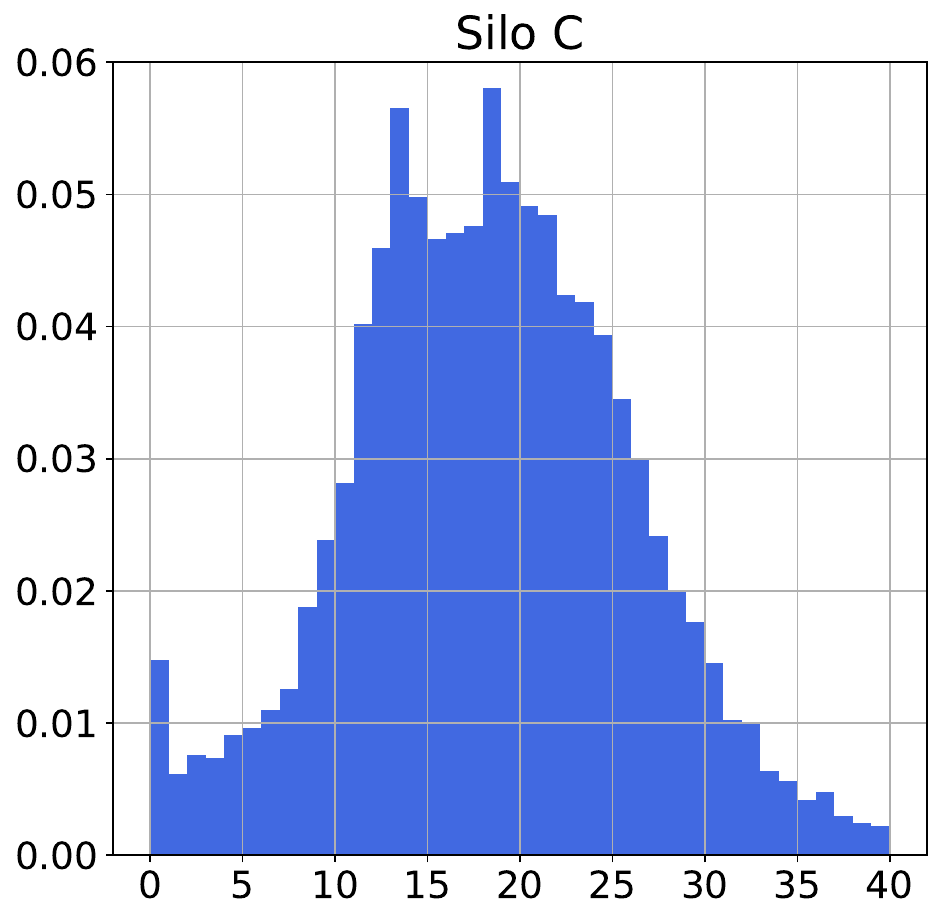}}
\caption{An example of probability distributions of \textit{$S^i_{m}$} and \textit{$S^j_{m}$}, where number of samples = 10,000 and \textit{m} denotes the number of discrete states. The x-axis denotes the indices of discrete states, while the y-axis indicates the probability of each state being visited.}
\label{fig:im_sampling}
\end{figure}

Once the transfer learning process starts, we encounter the second challenge of determining the relevance of the knowledge that can be transferred between players. This relevance is quantified by the weight parameter $\alpha_{\text{TF}}$. In state-based distributed manufacturing systems, we can employ an importance-weighted update procedure based on importance sampling via Jensen-Shannon divergence~\citep{Nielsen2019}. Importance sampling assesses the statistical similarity between two probability distributions of the players using Kullback-Leibler (KL) divergence~\citep{Kullback1997, MacKay2003}. In our context, Jensen-Shannon divergence measures the similarity between the visited states of the players in the environment, which may differ among players. This ensures that transfer is considered irrelevant when players find themselves in completely dissimilar states of the environment. To estimate the similarity between two players (e.g., player $i$ and $j$) based on state-visitation frequencies, we compute the expected value of the logarithmic difference between two discrete probability distributions, as illustrated in Fig.~\ref{fig:im_sampling}.

After getting the discrete probability distribution among players, we calculate the Jensen-Shannon divergence to obtain a smoothed version of KL divergence, which is defined as follows:
\begin{align}\label{eq:jsdivergence1}
M=\frac{1}{2}\left(P_{S_i^m}+P_{S_j^m}\right),
\end{align}
\begin{align}\label{eq:jsdivergence2}
JSD\left(P_{S_i^m}||P_{S_j^m}\right)=\frac{1}{2}D_{KL}\left(P_{S_i^m}||M\right)+\frac{1}{2}D_{KL}\left(P_{S_j^m}||M\right),
\end{align}
where $P_{S_i^m}$ denotes the probability distribution $P$ of player $i$ for state $m$, $M$ denotes the maximum number of discretized states, the expression $D_{KL}\left(q_1||q_2\right)$ refers to the KL divergence, a metric quantifying the difference between two probability distributions, which are $q_1$ as the target distribution and $q_2$ as a reference distribution.

To combine both findings for determining the initiation and weight of knowledge transfer, the weight parameter $\alpha_{\text{TF}}$ is computed based on the Jensen-Shannon divergence value of states $JSD\left(P_{S_i^m}||P_{S_j^m}\right)$ as well as the exploration rate $\epsilon$, under the following conditions:
\begin{equation} \label{eq:alphatf}
 \alpha_{TF}=\left\{
    \begin{array}{ll}
      0 \quad\quad\quad \text{   , if } \left(\epsilon \geq \beta_{TF}\right) \vee \left(\sum_{m=0}^M JSD\left(P_{S_i^m}||P_{S_j^m}\right) \geq 1\right);\\
      1 - \sum_{m=0}^M JSD\left(P_{S_i^m}||P_{S_j^m}\right) \text{   , otherwise};\\
    \end{array}
  \right.
\end{equation}

\subsection{TL-SbPGs without pre-defined similarities}\label{sec:similar}

In this subsection, we examine a scenario where knowledge regarding the similarity between players is not predefined. In real-world manufacturing systems, it is common for the similarity between actuators to be unknown or challenging to determine due to the large scale of the system. As a result, previous approaches such as the SW and MOM approaches become impractical when applied as standalone methods. Hence, we relax the assumption made in the previous subsection and establish conditions for inferring the similarity between players during their learning progress. This measure of similarity is then incorporated into the learning process by adjusting the utility function design and action updates accordingly.

In general, similarities can be defined based on policies, which outline the actions to be taken in the state-action space, or utilities obtained during training. We are particularly interested in low-dimensional representations of policies and utilities, from which a suitable similarity measure can be derived. To achieve this, we propose an approach for measuring similarity for such low-dimensional latent representations, namely using radial basis function (RBF) networks. To assess the similarity between two or more players, we suggest comparing the latent representations between them. In this subsection, we explain how these similarity measures can be integrated into the transfer learning frameworks.

\subsubsection{RBF networks}\label{sec:rbf_app}

We propose utilizing simple RBF networks with fixed mean and variance to model and represent the relationships within the state-action and state-utility space for each player. RBF networks~\citep{Buhmann2000} are a class of artificial neural networks commonly used for function approximation, classification, and interpolation. They are particularly effective in capturing local patterns in the data due to their localized response to inputs. The output of an RBF network is determined by the distance between the input and a set of fixed centres (means), which makes them well-suited for measuring similarity in low-dimensional latent spaces.

In our approach, we employ a simple RBF network with fixed mean $\mu_i$ and covariance $\Sigma_i$ to represent the player's state-action and state-utility space. Specifically, the RBF network is designed as a two-dimensional function for player $i$, defined by the following equation:
\begin{align}\label{eq:rbf}
\xi_i = [a_i \  U_i]^T = \sum_j \theta_{i,j}\frac{e^{(s_i - \mu_i)^T\Sigma^{-1}_i(s_i - \mu_i)}}{\sum_j e^{(s_j - \mu_j)^T\Sigma^{-1}_j(s_j - \mu_j)}},
\end{align}
where mean $\mu_i \in [0,1]^d$ and covariance matrix $\Sigma_i \in \mathbb{R}$ are fixed beforehand, which simplifies the network structure by reducing the complexity and computational cost. $d$ is the dimension of the state space. The remaining parameters $\theta_{i,j}$ with $j=1,\ldots,J$ constitute the $J$-dimensional latent space. The parameters $\theta_{i,j}$ can be fitted to the state-action and state-utility space by using least squares optimization, which minimizes the difference between the network’s predicted output and the true values in the state-action or state-utility space. To reduce the computational requirements and noise due to exploration, we update the latent space representation at regular intervals, specifically after every $H-$th step. In addition, we implement the gradient descent method to optimize parameters $\theta_{i,j}$ by taking the steepest descent direction at each iteration.

In the context of the proposed TL-SbPGs, we utilize RBF networks to measure the similarity between players by comparing their network representations, which helps determine which players are suitable for the transfer learning process. The structure defined in Eq.~\eqref{eq:rbf} allows the RBF network to represent local features of the state-action and state-utility spaces in a smooth and continuous manner, which makes it a robust approach for capturing relationships between players in the transfer learning process. Additionally, RBF networks can serve as an alternative to the MOM and SW approaches, which function as a transfer learning framework between players. Both of these points are discussed in detail in the next sub-subsection.

\subsubsection{Similarity measures implementations}\label{sec:simi_app}

As previously discussed, player similarities can be evaluated using trained RBF networks by analysing their respective latent space representations. This involves computing the similarity between player $n$ and target player $m$, denoted as $L_{n,m}$, through a basic squared loss method applied to the weight parameters of player $\theta_n$ and target player $\theta_m$, as demonstrated:
\begin{align}\label{eq:simloss}
L_{n,m}=\left(\theta^n_{i,j,t-1}-\theta^m_{i,j,t-1}\right)^2.
\end{align}
After obtaining the similarities between players, we have two options to proceed, such as (a) continuing with the previous approaches, either the SW or MOM approaches by incorporating the obtained similarities, and (2) continuing directly with RBF networks-based transfer without employing the SW and MOM approaches.

In the SW and MOM approaches, we operate under the assumption that player similarities are predefined, either based on prior knowledge or determined using Eq.~\eqref{eq:simloss}. When considering two players, $n$ and $m$, a relatively low value of $L_{n,m}$ indicates similarity between them. In such cases, transfer learning can take place, and the original utility function described in Eq.~\eqref{eq:newgame} can be re-employed.

Alternatively, we can directly incorporate the trained RBF networks into the transfer learning frameworks without the SW and MOM approaches. This involves storing the resulting similarity measures for a specific time horizon, $S^{\text{SM}}_{i,n,j,t}$. Subsequently, we substitute the auxiliary utility function in Eq.~\eqref{eq:swloss} for SW and~\eqref{eq:momloss} for MOM with the latent variables, which results in:
\begin{align}\label{eq:latentloss}
S^{\text{SM}}_{i,n,j,t} = \sum_{k=0}^{H-1} (\theta_{i,j,t-k} - \theta_{n,j,t-k})^2,
\end{align}

The transfer learning frameworks integrate the RBF-based similarity measures by updating the original utility function described in Eq.~\eqref{eq:newgame}, as given:
\begin{align}\label{eq:utility_latent_loss}
\tilde{U}_i(a_i, S^{A_i}) = U_i(a_i, S^{A_i}) \!-\! \frac{1}{N-1} \cdot \sum_{n=0}^{N}\alpha_{\text{TF}_{i,n}}S^{\text{SM}}_{i,n,j,t}.
\end{align}

\subsection{Update mechanism for TL-SbPGs}\label{sec:update} 

In this subsection, our focus is on demonstrating the integration of the proposed transfer learning schemes into SbPGs. Pseudocode~\ref{alg:tf} outlines the update loop of the training process for the proposed TL-SbPG methods, which considers both predefined and undefined similarities between players. Furthermore, the players employ a TL-SbPG learning algorithm with a global interpolation technique~\citep{Schwung2020} to estimate the value of actions in a specific state, as well as Best Response learning~\citep{Schwung2020} as their policies. The selected actions are then communicated to the environment by the players.
\begin{algorithm}
\caption{Basic approach of TL-SbPGs.}\label{alg:tf}
\KwData{Actual Episode, Maximum Episode, Duration per Episode, Position of Support Vectors}
\KwResult{Next Action, Performance Maps}
initialize all hyperparameters\;
\If{SW or MOM approaches}{select players that are involved in the TL\;}
\While{maximum episode has not reached}{
    \While{maximum time has not reached}{
        calculate and discretize current states\;
        \If{update action}{
            store visited states for the TL players\;
            calculate $\alpha_{\text{TF}}$\;
            \eIf{SW or MOM approaches}{
            calculate loss function, $H^{\text{SW}}_{i,j}(a_i, a_j)$ or $H^{\text{MOM}}_{i,j,t}(a_i, a_j)$\;}{
            compute current latent representations, $\theta_{i,j,t}$ and $\theta_{n,j,t}$\;
            calculate loss function, $S^{\text{SM}}_{i,n,j,t}$\;
            optimize networks\;}
            calculate utility, $\tilde{U}_i(a_i, S^{A_i})$\;
            update the performance map\;
            \eIf{exploring}{
                explore a random action\;}{
                globally interpolate an action from the performance map\;}
            }
        }
    }
\end{algorithm}

With TL-SbPG methods established, we evaluate their effectiveness in lab and scalable testbeds, comparing them to baselines across various learning scenarios.

\section{Results and discussions}\label{sec:results}

In this study, we evaluate the effectiveness of the proposed TL-SbPG methods by applying them to two laboratory testbeds, namely the bulk good system (BGS) and the larger-scale BGS (LS-BGS). The testbeds and their learning setup are defined in Sections~\ref{sec:testbed} and~\ref{sec:scenario}, respectively.

Moreover, we conduct three experiments in the subsequent subsections, such as:
\begin{itemize}
    \item TL-SbPGs with pre-defined similarities applied to the BGS
    \item TL-SbPGs without pre-defined similarities applied to the BGS
    \item TL-SbPGs with a combination of both approaches (as explained in Section~\ref{sec:simi_app}) applied to the LS-BGS
\end{itemize}
We also compare the results of each comparison to the baseline algorithm, which is the vanilla SbPGs. Additionally, we perform several ablation studies to validate the scalability and flexibility of the TL-SbPGs by examining their plug-and-play capabilities on the testbed. These ablation studies also evaluate whether the number of players involved in the transfer learning process influences performance, as well as assess the impact of the threshold $\beta_{TF}$ on the MOM and SW approaches.

\subsection{Laboratory testbed}\label{sec:testbed}

\begin{figure}[ht]
 \centering
 \includegraphics[width=0.95\columnwidth,keepaspectratio]{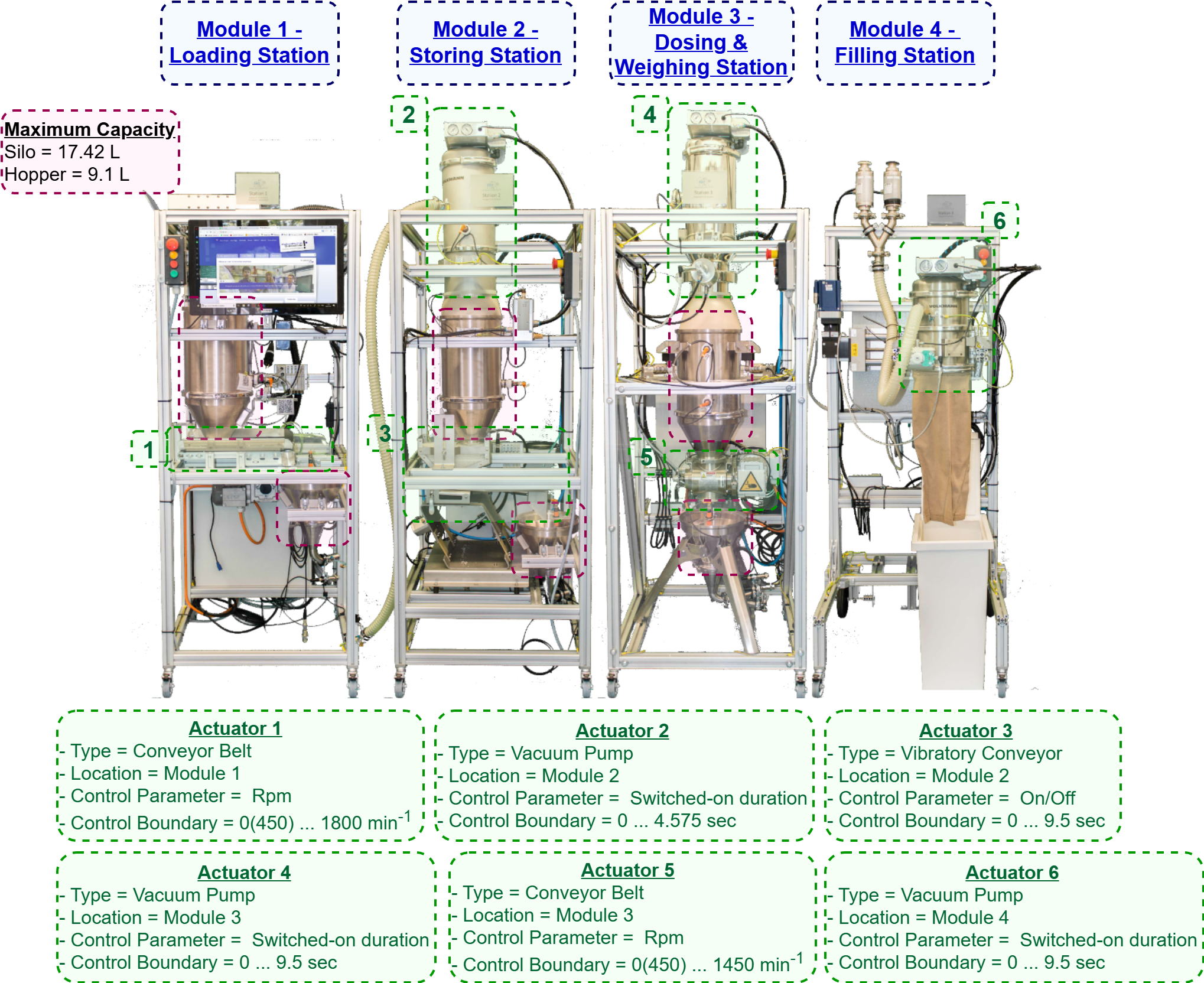}
\caption{Schematic of the laboratory scale testbed, namely the BGS.}
\label{fig:testbed}
\end{figure}

The BGS~\citep{Schwung2020} represents a flexible and smart production system designed to handle material transportation, with a focus on bulk materials. This approach can also be extended to energy systems, where similar decentralized optimization principles can be applied. The methodologies discussed in~\citep{Schwung2020b} demonstrate how these concepts can be effectively translated to energy applications, which further highlights their industrial relevance. The laboratory-scale testbed operates within four modules, as illustrated in Fig.~\ref{fig:testbed}. Module~1 and Module~2 serve as typical supply, buffer, and transportation stations, which are controlled by a continuously operated belt conveyor in Module~1 and by a vacuum pump and a vibratory conveyor in Module~2. Module~3 is the dosing station, which is controlled by a second vacuum pump and a continuously operated rotary feeder. Then, Module~4 is the filling and final station, which operates to fill up the transport containers.

Each station contains one or more reservoirs for temporarily storing transported materials, where silos have a maximum capacity of 17.42L and hoppers hold up to 9.1L. Different actuators control the flow of materials into and out of the reservoirs. For example, in Module 1, the actuator is a conveyor belt with a motor speed control parameter ranging from 0 to 1800 rpm, while the vacuum pumps in each module are controlled by the duration of their operation, with varying control ranges. Additionally, in Module 2, the vibratory conveyor is controlled by simply switching it fully on or off. The details of the reservoirs and actuators are shown in Fig.~\ref{fig:testbed}. Furthermore, each module can be flexibly positioned in different production sequences, or added and removed entirely, thanks to its plug-and-play functionality.

Each module owns an individual PLC-based control system using a Siemens ET200SP and several sensors to monitor the modules' state and enable control of each module. The communication between the modules is developed via Profinet. Moreover, an RFID reader and an RFID tag are assigned in each module to identify the subsequent module in the production flow. The BGS simulation is available on an open-source machine learning framework, namely MLPro~\citep{Detlef2022a, Detlef2022b}. In this research, we initially utilize the default configuration of the BGS. For the ablation studies, we then test different setups of the BGS by using its plug-and-play capability, as well as evaluate the effects of varying control parameters of the actuators.

\begin{figure}[ht]
 \centering
 \includegraphics[width=1.00\columnwidth,keepaspectratio]{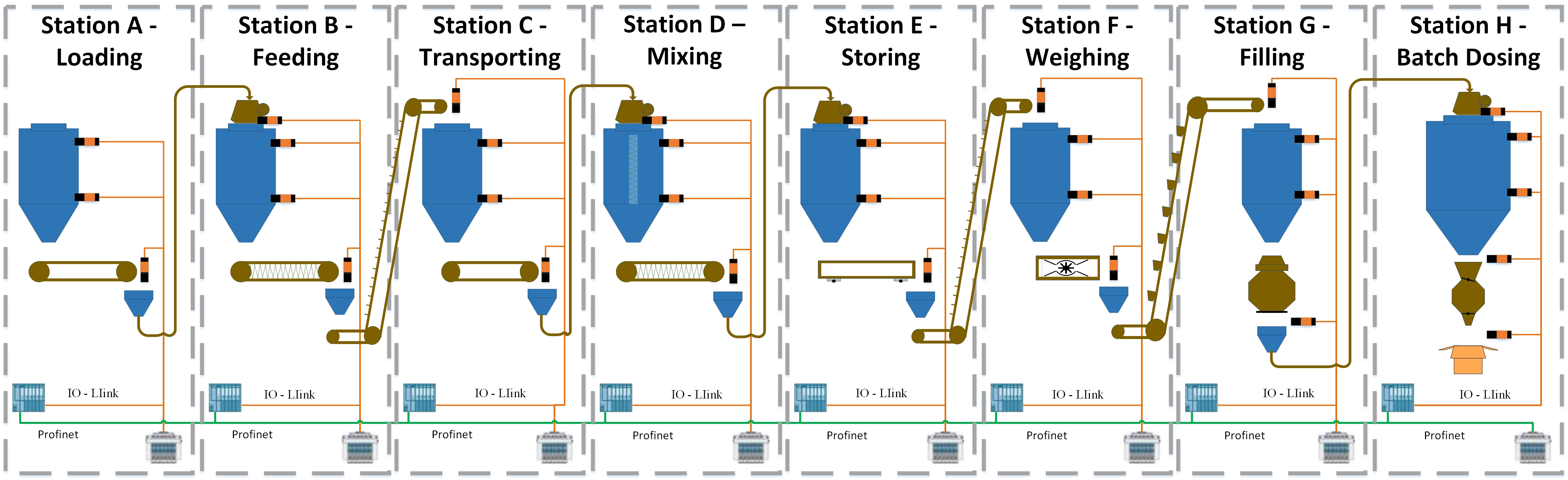}
\caption{Schematic of the LS-BGS~\citep{Yuwono2023b}.}
\label{fig:ls-bglp}
\end{figure}

Furthermore, we have extended the BGS into a larger-scale production scenario in our previous research~\citep{Yuwono2023b} that allows different settings, including serial-parallel processes. In this research, we utilize first the LS-BGS in a sequential setting with 8 modules and 15 actuators as the second test environment, which is shown in Fig.~\ref{fig:ls-bglp}. Next, we also test serial-parallel processes within this environment to validate the robustness and scalability of the proposed approach. In parallel, this demonstrates that the LS-BGS handles more complex systems while maintaining highly flexible configurations. The actuators and reservoirs have more variety and a wider range of parameter controls that represent the real industrial processes. Moreover, the LS-BGS simulation is available in MLPro-MPPS~\citep{Yuwono2023c, Yuwono2023}.

Before presenting experiment results, we outline the learning setup, including system configurations, training procedures, and utility formulations used in our experiments.

\subsection{Learning setup}\label{sec:scenario}

In this subsection, we explain how the operation scenario of the BGS is incorporated with TL-SbPGs. First, we consider a continuous production scenario in which the process flow has a constant cycle with continuous throughput. The main target is to fulfil a specified production demand in volume over time while reducing power consumption as much as possible and avoiding overflow. Therefore, such an operation scenario leads us to a multi-objective optimization. PG is being arranged within the BGS environment, in which each actuator serves as a player and the fill level indicators of reservoirs serve as state variables, see more details in~\citep{Schwung2020}.

Additionally, we apply the identical actions, states, and utility functions in TL-SbPGs and the baseline (vanilla SbPGs) to ensure that the results are comparable. The local utility function for each player $i$ is defined as:
\begin{equation} \label{Equation:4-1_UtilityFunction}
 U_i=\left\{
    \begin{array}{ll}
    \!\frac{1}{1\!+\!\alpha_{L}L_{p}^{i}}\!+\!\frac{1}{1\!+\!\alpha_{P}P^{i}}\!+\!\frac{1}{1\!-\!\alpha_{D}V_{D}}\!\ \text{    , if } i=N;\\
    \!\frac{1}{1\!+\!\alpha_{L}L_{p}^{i}}\!+\!\frac{1}{1\!+\!\alpha_{L}L_{s}^{i}}\!+\!\frac{1}{1\!+\!\alpha_{P}P^{i}}\!\quad\text{, otherwise};\\
    \end{array}
  \right.
\end{equation}
where $P^i$ denotes the power consumption in the current iteration, $L_{p}^{i}$ and $L_{s}^{i}$ are local constraints to prevent overflow and emptiness of the prior and subsequent reservoir respectively, $V_{D}$ denotes the production demand objective, $\alpha_L$, $\alpha_D$, $\alpha_P$ are predefined weighing parameters, and $i$ identifies the player, i.e. $i=N$ means the last player. Moreover, Eq.~\eqref{Equation:4-2-1_DemandVolume} and Eq.~\eqref{Equation:4-2-1_Badewanne} describe the computation of $V_{D}$,  $L_{p}^{i}$ and $L_{s}^{i}$.
\begin{align}\label{Equation:4-2-1_DemandVolume}
V_{D}=\smashoperator{\int_{0}^{T_{I}}}\dot{D}_t\,dt,  \quad \dot{D}_t=
\begin{dcases}
\dot{V}_{N,out}-\dot{V}_{N,in} &\text{, if}\ h^{N}=0\;\\
0&\text{, otherwise};\\
\end{dcases}
\end{align}
where $T_I$ is the iteration length and $\dot{V}_{N,out}$, $\dot{V}_{N,in}$ are the outflow and inflow to the module. $h^{N}$ denotes the fill level of the related buffer that has been normalized. We remark that the last player applies the demand object exclusively, where the demand cannot be satisfied while the hopper of the last station has a lower fill level than the desired demand.

\begin{equation} \label{Equation:4-2-1_Badewanne}
L_{p}^{i}=\smashoperator{\int_{0}^{T_{I}}}\mathbbm{1}_{h^i_{p}<H^i_{p}}(h^i_{p})\,dt, \quad L_{s}^{i}=\smashoperator{\int_{0}^{T_{I}}}\mathbbm{1}_{h^i_{s}>H^i_{s}}(h^i_{s})\,dt, 
\end{equation}
where $H^i_{p}$ and $H^i_{s}$ are lower limit and upper limit for the fill-levels. The main role of $L_{p}^{i}$ and $L_{s}^{i}$ is to avoid overflow and a point of congestion or a bottleneck situation.  

Concerning the LS-BGS environment, we encountered a challenge in achieving the production demand while using Eq.~\eqref{Equation:4-1_UtilityFunction} as the utility function. Despite observing an increase in the potential value, the production demand could not be consistently met. After careful analysis, we identified the cause of the issue, which is that only the last player considered the production demand in their utility function, while the other 14 players did not. As a result, the players tended to prioritize consolidation rather than collectively striving to fulfill the production demand. To address this problem and account for the number of modules involved, we propose incorporating the production demand objective $V_{D}$ into each module's utility function. By doing so, we ensure that the utility function aligns with the concept of SbPGs and encourages all players to actively contribute towards meeting the production demand. Hence, the local utility function for each player $i$ is defined as follows,
\begin{equation} \label{Equation:4-2_UtilityFunction_ls}
 U_i=\left\{
    \begin{array}{ll}
    \!\frac{1}{1\!+\!\alpha_{L}L_{p}^{i}}\!+\!\frac{1}{1\!+\!\alpha_{P}P^{i}}\!+\!\frac{1}{1\!-\!\alpha_{D}V_{D}}\!\ \quad\quad\quad\quad \text{   , if } i=N;\\
    \!\frac{1}{1\!+\!\alpha_{L}L_{p}^{i}}\!+\!\frac{1}{1\!+\!\alpha_{L}L_{s}^{i}}\!+\!\frac{1}{1\!+\!\alpha_{P}P^{i}}\!+\!\frac{1}{1\!-\!\alpha_{D}V_{D}}\! \quad \text{, otherwise};\\
    \end{array}
  \right.
\end{equation}

We train the players for 9 training episodes with $1\cdot10^6$ iterations for each episode. Each iteration corresponds to 1 second of operation on the real machine. The policy adaptation takes place every 10 iterations, and each state information is discretized into 40 equidistant spaces. We set the production demand at a constant rate of 0.125 L/s. To ensure efficient computation and representation, we limit the dimension of the latent space representation to 3x3 for RBF networks.

With the experimental setup defined, we begin our analysis with results from the BGS.

\subsection{Results on the BGS}\label{sec:res_3}

In the BGS environment, the selection of players for transfer learning is defined based on the types and operation parameters of the actuators involved. One of our main objectives in this multi-objective problem is to optimize power consumption. To identify the actuators with the highest power consumption, we measure the proportion of total power consumed by each actuator during normal operation, as presented in Table~\ref{Tab:energy}. Upon analyzing the power consumption distribution, we observe that more than half of the total power consumption is attributed to vacuum pumps. Consequently, we decided to incorporate these two actuators in the transfer learning process because, at the same time, both of them have nearly identical characteristics. By optimizing the performance of these critical components, we aim to achieve significant improvements in power efficiency within the BGS system.
\begin{table}[ht]
\renewcommand{\arraystretch}{1.0}
\caption{The proportion of power consumption by each actuator during normal operation.}
\label{Tab:energy}
\centering
  \begin{tabular}{c||c|c}
\hline
    \bfseries No. & \bfseries Actuator/Player & \bfseries Power Consumption\\ \hline \hline
  1& Belt Conveyor A & 9.7\%\\ \hline
  2& Vacuum Pump B & 17.5\%\\ \hline
  3& Vibratory Conveyor B & 6.3\%\\ \hline
	4& Vacuum Pump C & 35.7\%\\ \hline
  5& Rotary Feeder C & 30.9\%\\ \hline
	  \hline
 \end{tabular}
\end{table}

Through our experiments, we found that the TL-SbPG methods achieved comparable performance to the vanilla SbPG, but within a significantly shorter time frame, which requires approximately 2-4 fewer training episodes (or 22.2-44.4\% less) to converge. Additionally, we observed that as the training continued uninterrupted for the full 9 training episodes, the performance of the transfer learning methods improved even further. We claim that the players could explore more state-action combinations. However, we note that the standard SbPG approach provided an optimal policy with respect to avoiding overflow and meeting production demands, and thus, the power consumption was the primary focus for achieving convergence with reduced consumption.

\begin{figure}[t]
 \centering
 \subfloat[Power consumption over time (x-axis: simulation time in 100s, y-axis: power in kW).]{\includegraphics[width=\columnwidth,keepaspectratio]{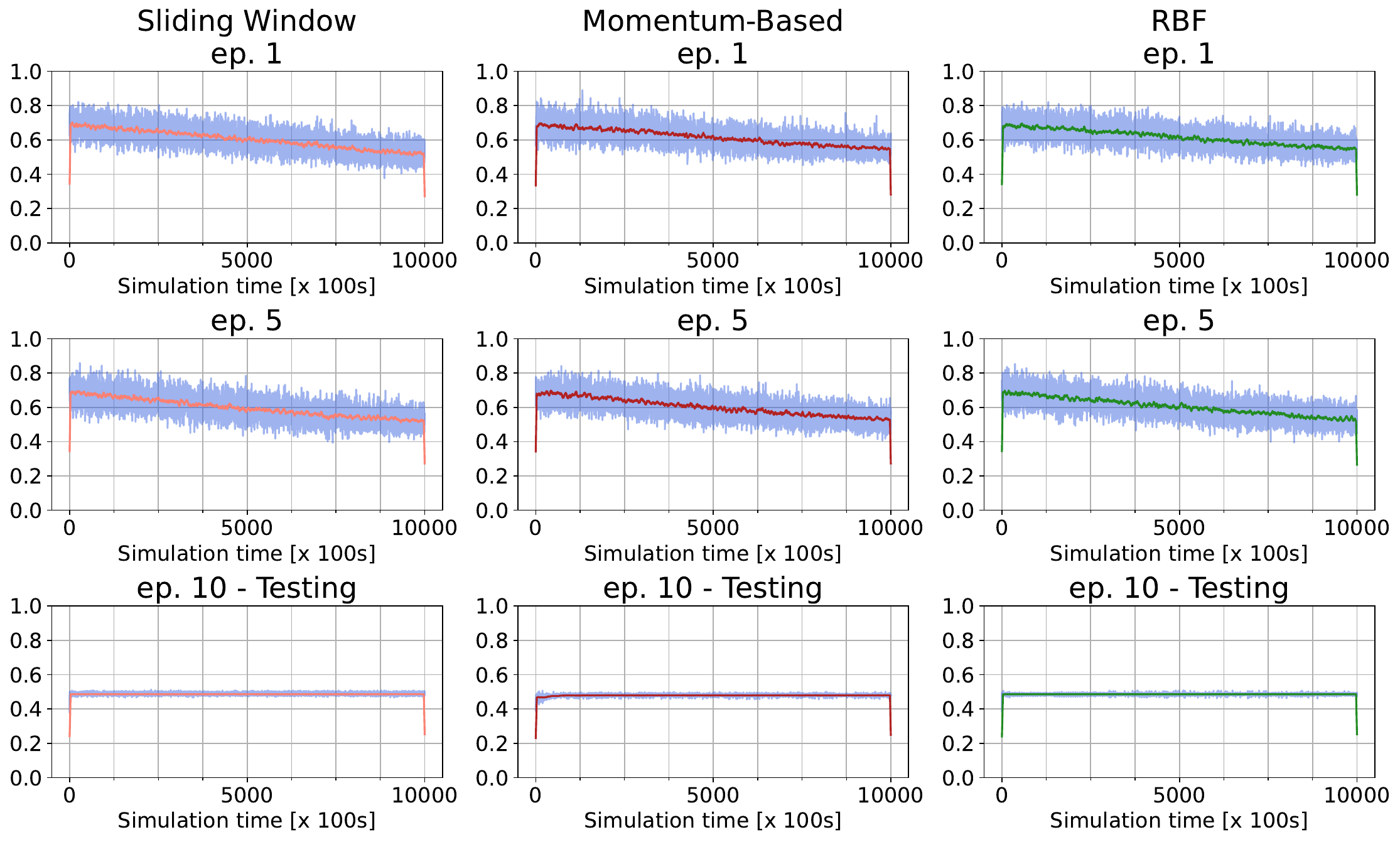}} 
  \qquad 
 \subfloat[Potential value over time (x-axis: simulation time in 100s, y-axis: potential value).]{\includegraphics[width=\columnwidth,keepaspectratio]{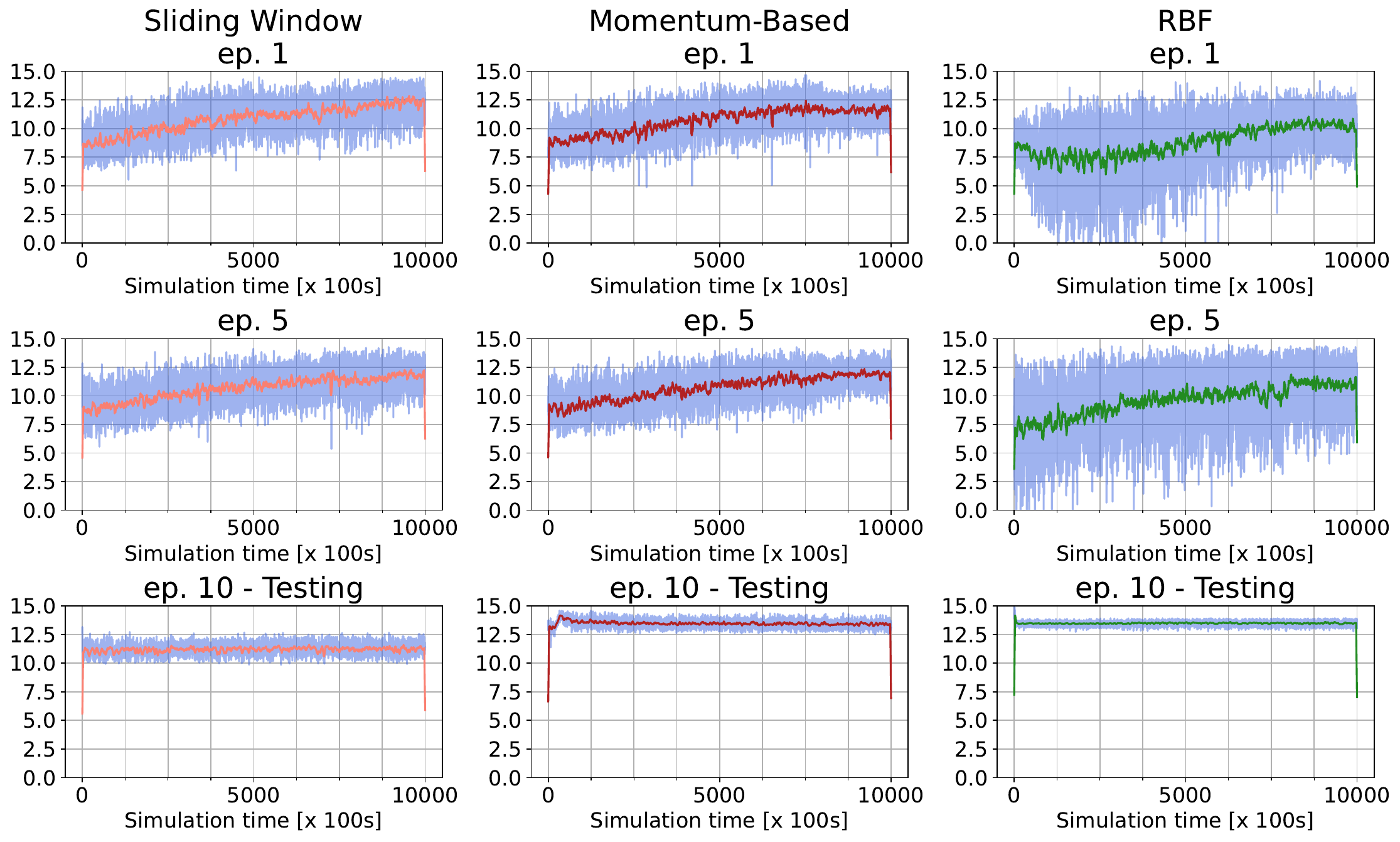}}
\caption{Power consumption and potential value of specific episodes using various TL-SbPG approaches in the BGS.}
\label{fig:res_bglp}
\end{figure}

Fig.~\ref{fig:res_bglp} illustrates the power consumption and potential value during the training and testing processes. The graph demonstrates a consistent and significant reduction in power consumption over time across each method. This reduction in power consumption has a direct impact on the global multi-objective of the production system, as outlined in Eq.~\eqref{Equation:4-1_UtilityFunction}. By achieving a notable decrease in power consumption, the overall performance and efficiency of the production system, as well as the potential value are improved, aligning with the system's objectives.

\begin{figure}[ht]
 \centering
\subfloat[SW approach, x-axis: fill-level of previous buffer (\%), y-axis: fill-level of subsequent buffer (\%).]{\includegraphics[width=0.95\columnwidth,keepaspectratio]{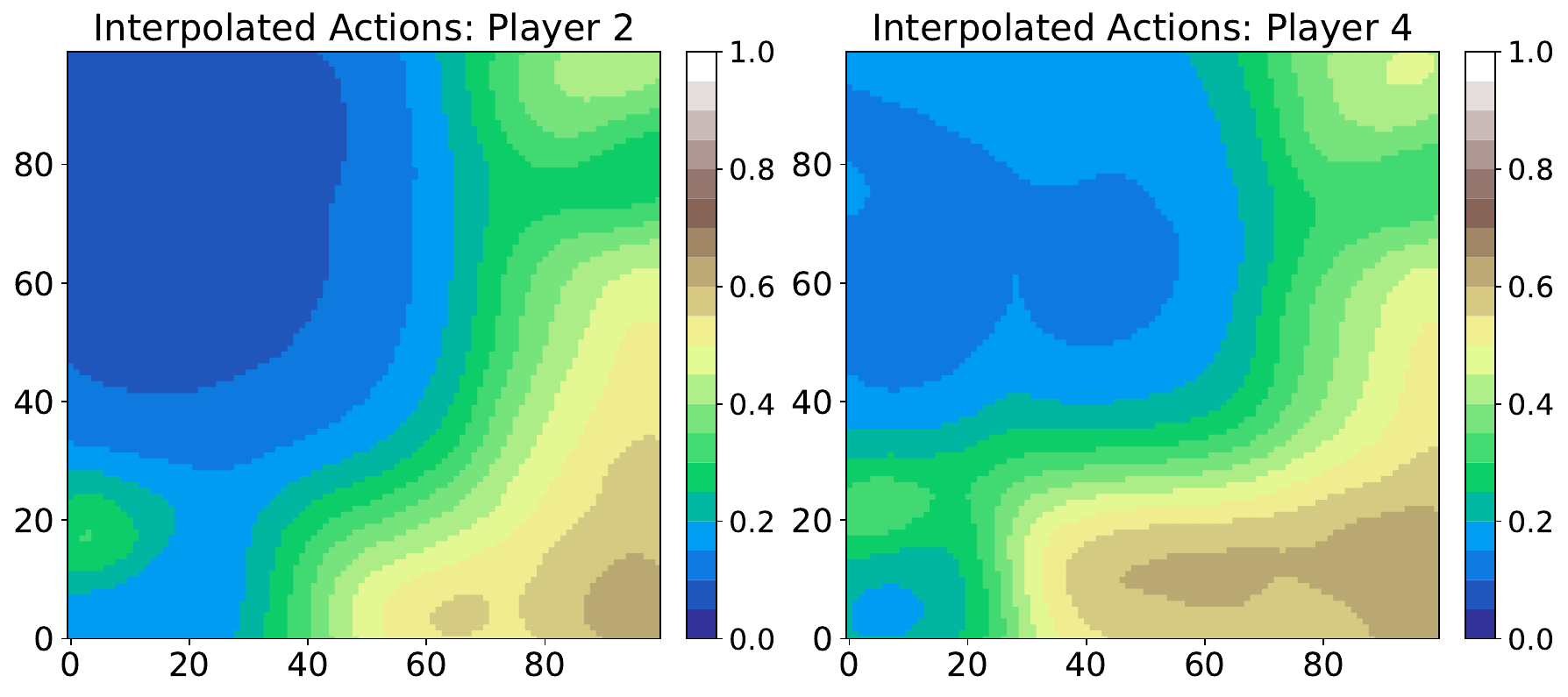}} 
  \qquad 
 \subfloat[MOM approach, x-axis: fill-level of previous buffer (\%), y-axis: fill-level of subsequent buffer (\%).]{\includegraphics[width=0.95\columnwidth,keepaspectratio]{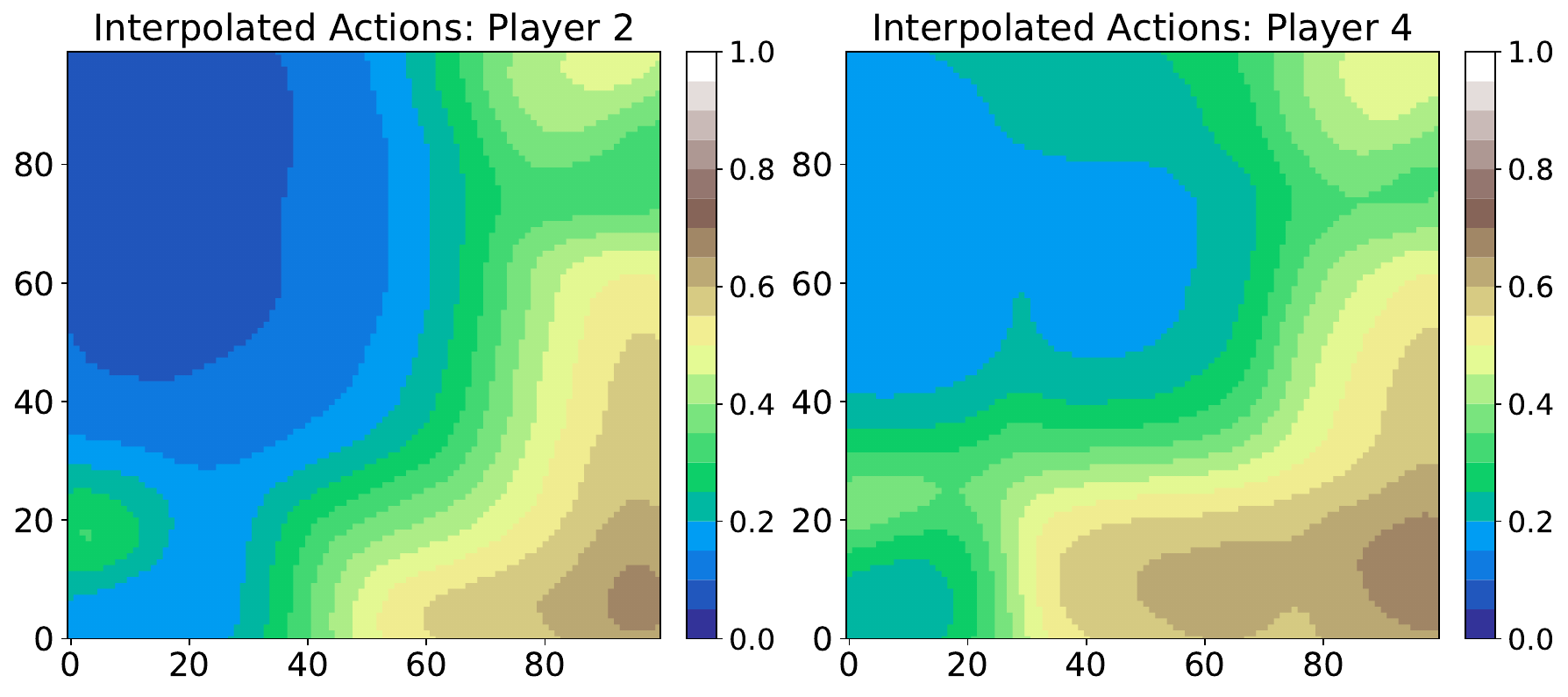}}
\caption{Performance maps of Players 2 and 4.}
\label{fig:heatmaps_res}
\end{figure}

Fig.~\ref{fig:heatmaps_res} displays the performance maps of two involved actuators, namely Player 2 (vacuum pump in station B) and Player 4 (vacuum pump in station C), using the SW and MOM approaches, respectively. The x-axis represents the fill level of prior reservoirs, while the y-axis represents the fill level of the next reservoirs. By analyzing the heat maps, we observe that the performance maps for both vacuum pumps are very similar because of the transfer learning. In conditions where the fill level of the prior reservoir is high and the next reservoir is low, the vacuum pumps remain activated for a longer duration. This indicates that vacuum pumps are effectively utilized to prevent overflow and maintain a smooth flow of materials. Conversely, in conditions where the prior reservoir is low and the next reservoir is high, the vacuum pumps are less activated to avoid potential overflow.

\begin{figure}[ht]
 \centering
 \includegraphics[width=1.0\columnwidth,keepaspectratio]{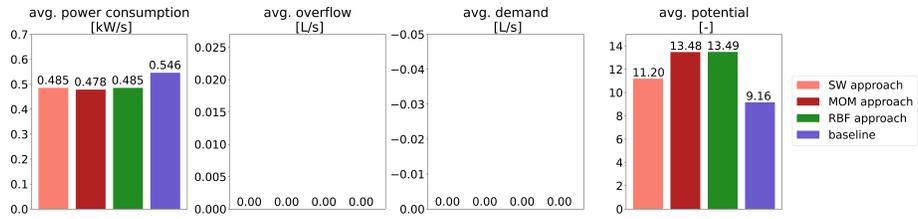}
\caption{Comparison between SbPGs and TL-SbPGs on the BGS' performances.}
\label{fig:comparison_bglp}
\end{figure}

In Fig.~\ref{fig:comparison_bglp}, we present a comparison of the overall performances between each TL-SbPG approach and the baseline. The results show that the TL-SbPG approaches outperform the baseline in two aspects. First, the power consumption is significantly reduced, ranging up to 12.46\%. Second, the potential value is enhanced by approximately between 22.3\% and 47.2\% compared to the baseline.

Additionally, we tune and identify the optimal hyperparameter configuration using an automated multi-resolution grid search~\citep{Bergstra2013} to guarantee the optimal configuration of the learning algorithms. One of the hyperparameters is the threshold $\beta_{\text{TF}}$, which plays a crucial role as a kick-off indicator of the transfer learning process, as explained in Eq.~\eqref{eq:alphatf}. For the TL-SbPG with pre-defined similarities, the best results are obtained when the threshold $\beta_{\text{TF}}$ for the MOM and SW approaches are 0.80 and 0.40, respectively. Meanwhile, for the TL-SbPG without pre-defined similarities, the threshold $\beta_{\text{TF}}$ for RBF networks is 0.24. Table~\ref{tab:ablation_tf} shows an ablation study of threshold $\beta_{\text{TF}}$ on the MOM and SW approaches.

\begin{table}[ht]
\renewcommand{\arraystretch}{1.0}
\caption{Ablation study of threshold $\beta_{\text{TF}}$.}
\label{tab:ablation_tf}
\centering
\begin{tabular}{|c||c|c|c|c|}
\hline
\multicolumn{1}{|c||}{\textbf{$\beta_{\text{TF}}$}} & \multicolumn{1}{c|}{\textbf{\begin{tabular}[c]{@{}c@{}}Avg. Power\\ {[}kw/s{]}\end{tabular}}} & \multicolumn{1}{c|}{\textbf{\begin{tabular}[c]{@{}c@{}}Avg. Overflow\\ {[}L/s{]}\end{tabular}}} & \multicolumn{1}{c|}{\textbf{\begin{tabular}[c]{@{}c@{}}Avg. Demand\\ {[}L/s{]}\end{tabular}}} & \textbf{\begin{tabular}[c]{@{}c@{}}Potential\\ {[}-{]}\end{tabular}} \\ \hline\hline
\multicolumn{5}{|c|}{\textbf{SW approach, where H = 10}}                                                                                                                                                                                                                                                                                                                                                                         \\ \hline\hline
\multicolumn{1}{|c||}{0.2}              & \multicolumn{1}{c|}{0.476}                                                                                & \multicolumn{1}{c|}{0.000}                                                                      & \multicolumn{1}{c|}{-0.007}                                                                   & 9.953                                                                     \\ \hline
\multicolumn{1}{|c||}{0.4}              & \multicolumn{1}{c|}{0.485}                                                                                & \multicolumn{1}{c|}{0.000}                                                                      & \multicolumn{1}{c|}{0.000}                                                                    & 11.209                                                                    \\ \hline
\multicolumn{1}{|c||}{0.6}              & \multicolumn{1}{c|}{0.511}                                                                                & \multicolumn{1}{c|}{0.415}                                                                      & \multicolumn{1}{c|}{0.000}                                                                    & 8.775                                                                     \\ \hline
\multicolumn{1}{|c||}{0.8}              & \multicolumn{1}{c|}{0.465}                                                                                & \multicolumn{1}{c|}{0.040}                                                                      & \multicolumn{1}{c|}{0.000}                                                                    & 10.984                                                                    \\ \hline\hline
\multicolumn{5}{|c|}{\textbf{MOM approach, where $\alpha_{MOM}$ = 0.50}}                                                                                                                                                                                                                                                                                                                                                  \\ \hline\hline
\multicolumn{1}{|c||}{0.2}              & \multicolumn{1}{c|}{0.485}                                                                                & \multicolumn{1}{c|}{0.000}                                                                      & \multicolumn{1}{c|}{0.000}                                                                    & 10.650                                                                    \\ \hline
\multicolumn{1}{|c||}{0.4}              & \multicolumn{1}{c|}{0.482}                                                                                & \multicolumn{1}{c|}{0.515}                                                                      & \multicolumn{1}{c|}{0.000}                                                                    & 9.048                                                                     \\ \hline
\multicolumn{1}{|c||}{0.6}              & \multicolumn{1}{c|}{0.484}                                                                                & \multicolumn{1}{c|}{0.000}                                                                      & \multicolumn{1}{c|}{0.000}                                                                    & 12.145                                                                    \\ \hline
\multicolumn{1}{|c||}{0.8}              & \multicolumn{1}{c|}{0.478}                                                                                & \multicolumn{1}{c|}{0.000}                                                                      & \multicolumn{1}{c|}{0.000}                                                                    & 13.482                                                                    \\ \hline
\end{tabular}
\end{table}

Then, we validate the effectiveness of TL-SbPGs in arbitrary BGS environments, including serial-parallel processes and module additions. We follow the three distinct scenarios outlined in~\citep{Yuwono2023a}. Instead of training the policies from scratch, we reuse the pre-trained policies from the default BGS setup and apply an additional 1-2 training episodes, depending on the scenario's complexity. This approach, as demonstrated in~\citep{Schwung2020}, proves effective and highlights the system's plug-and-play functionality. In serial-parallel processes with the additional module configuration of 1-3-2//3-4, the transfer learning process shifts from involving both vacuum pumps to focusing on the actuators within the duplicated Module 3 because they are identical. Table~\ref{tab:ablation_bgs} provides a comparative summary of the TL-SbPG results across different scenarios, which shows that the proposed approaches consistently outperform the vanilla SbPG baseline, with MOM and RBF delivering the best performance compared to the SW approach.
\begin{table}[ht]
\renewcommand{\arraystretch}{1.0}
\caption{Comparison between SbPGs and TL-SbPGs on the arbitrary environments of the BGS.}
\label{tab:ablation_bgs}
\centering
\begin{tabular}{|ccccc|}
\hline
\multicolumn{1}{|c||}{\textbf{Algorithm}} & \multicolumn{1}{c|}{\textbf{\begin{tabular}[c]{@{}c@{}}Demand\\ (L/s)\end{tabular}}} & \multicolumn{1}{c|}{\textbf{\begin{tabular}[c]{@{}c@{}}Overflow\\ (L/s)\end{tabular}}} & \multicolumn{1}{c|}{\textbf{\begin{tabular}[c]{@{}c@{}}Power\\ (kW/s)\end{tabular}}} & \textbf{Potential} \\ \hline\hline
\multicolumn{5}{|c|}{Sequence: 1-3-2-4} \\ \hline\hline
\multicolumn{1}{|c||}{Baseline} & \multicolumn{1}{c|}{0.000} & \multicolumn{1}{c|}{0.000} & \multicolumn{1}{c|}{0.479} & 11.960 \\ \hline
\multicolumn{1}{|c||}{SW} & \multicolumn{1}{c|}{0.000} & \multicolumn{1}{c|}{0.000} & \multicolumn{1}{c|}{0.465} & 12.065 \\ \hline
\multicolumn{1}{|c||}{MOM} & \multicolumn{1}{c|}{0.000} & \multicolumn{1}{c|}{0.000} & \multicolumn{1}{c|}{0.460} & 12.897 \\ \hline
\multicolumn{1}{|c||}{RBF} & \multicolumn{1}{c|}{0.000} & \multicolumn{1}{c|}{0.000} & \multicolumn{1}{c|}{0.460} & 12.844 \\ \hline\hline
\multicolumn{5}{|c|}{Sequence: 1-2//3-4} \\ \hline\hline
\multicolumn{1}{|c||}{Baseline} & \multicolumn{1}{c|}{0.000} & \multicolumn{1}{c|}{0.000} & \multicolumn{1}{c|}{0.488} & 11.883 \\ \hline
\multicolumn{1}{|c||}{SW} & \multicolumn{1}{c|}{0.000} & \multicolumn{1}{c|}{0.000} & \multicolumn{1}{c|}{0.471} & 11.997 \\ \hline
\multicolumn{1}{|c||}{MOM} & \multicolumn{1}{c|}{0.000} & \multicolumn{1}{c|}{0.000} & \multicolumn{1}{c|}{0.465} & 12.765 \\ \hline
\multicolumn{1}{|c||}{RBF} & \multicolumn{1}{c|}{0.000} & \multicolumn{1}{c|}{0.000} & \multicolumn{1}{c|}{0.467} & 12.800 \\ \hline\hline
\multicolumn{5}{|c|}{Sequence: 1-3-2//3-4} \\ \hline\hline
\multicolumn{1}{|c||}{Baseline} & \multicolumn{1}{c|}{0.000} & \multicolumn{1}{c|}{0.012} & \multicolumn{1}{c|}{0.669} & 15.410 \\ \hline
\multicolumn{1}{|c||}{SW} & \multicolumn{1}{c|}{0.000} & \multicolumn{1}{c|}{0.008} & \multicolumn{1}{c|}{0.615} & 15.984 \\ \hline
\multicolumn{1}{|c||}{MOM} & \multicolumn{1}{c|}{0.000} & \multicolumn{1}{c|}{0.001} & \multicolumn{1}{c|}{0.603} & 17.010 \\ \hline
\multicolumn{1}{|c||}{RBF} & \multicolumn{1}{c|}{0.000} & \multicolumn{1}{c|}{0.002} & \multicolumn{1}{c|}{0.612} & 16.867 \\ \hline
\end{tabular}
\end{table}

We conducted a paired t-test to compare the utility values of TL-SbPGs and SbPGs across 12 experiments on the BGS under varied setups and TL-SbPGs variances. The results showed that TL-SbPGs significantly outperformed SbPGs, with a mean utility improvement of 1.51 ($t(11) = 3.70, p = 0.0035$). This confirms that TL-SbPGs provide consistently better performance under diverse conditions.

Finally, we modified the power consumption and control parameters of the vacuum pumps in the default setting of the BGS, representing Player 2 and Player 4, both involved in the transfer learning process. Specifically, we reduced Player 2's transport capability by 25\% and increased its power consumption by 15\%, while for Player 4, we increased the transport capability by 10\% and raised the power consumption by 25\%. Due to these changes in control parameters, unlike the previous ablation study, we retrained the system from scratch. The results, shown in Table~\ref{tab:ablation_control}, demonstrate that the proposed approach again significantly outperforms the baseline. The findings from the last two experiments confirm that TL-SbPG remains effective across varying environments and control parameters, further supporting the validity of our convergence proof and theorem in this study.
\begin{table}[ht]
\renewcommand{\arraystretch}{1.0}
\caption{Comparison between SbPGs and TL-SbPGs on the BGS' performances with modified control and power consumption parameters of Player 2 and 4.}
\label{tab:ablation_control}
\centering
\begin{tabular}{|c||c|c|c|c|}
\hline
\textbf{Algorithm} & \textbf{\begin{tabular}[c]{@{}c@{}}Demand\\ (L/s)\end{tabular}} & \textbf{\begin{tabular}[c]{@{}c@{}}Overflow\\ (L/s)\end{tabular}} & \textbf{\begin{tabular}[c]{@{}c@{}}Power\\ (kW/s)\end{tabular}} & \textbf{Potential} \\ \hline\hline
Baseline & 0.000 & 0.000 & 0.710 & 6.741 \\ \hline
SW & 0.000 & 0.000 & 0.629 & 8.784 \\ \hline
MOM & 0.000 & 0.000 & 0.618 & 8.901 \\ \hline
RBF & 0.000 & 0.000 & 0.621 & 8.775 \\ \hline
\end{tabular}
\end{table}

\subsection{Results on the LS-BGS}\label{sec:res_5}

\begin{figure}[t]
 \centering
 \includegraphics[width=0.70\columnwidth,keepaspectratio]{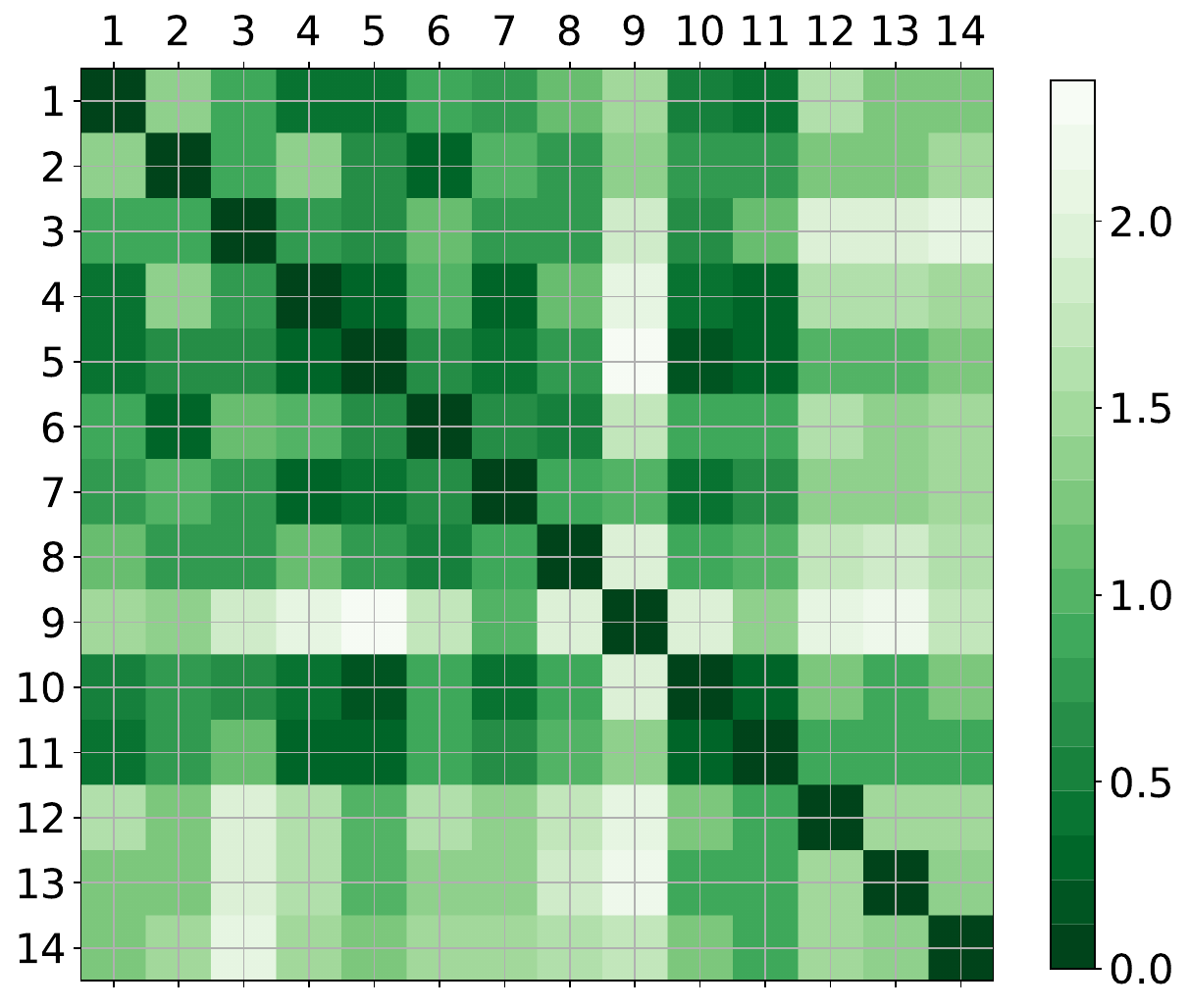}
\caption{Similarity matrix of actuators in the LS-BGS. The x- and y-axes represent player indices, and the color scale indicates the degree of similarity between each pair of players.}
\label{Fig:sim_matrix}
\end{figure}

To address the challenge of selecting which actuators to involve in transfer learning in the LS-BGS, we employ RBF networks to measure the similarity between the actuators as described in Eq.~\eqref{eq:simloss}. The resulting similarity matrix is depicted in Fig.~\ref{Fig:sim_matrix}. Based on the similarity matrix, we test different transfer learning scenarios involving different pairs of actuators and find that the optimal transfer learning settings involve three pairs of actuators: Players 5 and 10 (Conveyor Belt C and Belt Elevator F), Players 2 and 6 (Vacuum Pump B and Vacuum Pump D), as well as Players 1 and 4 (Conveyor Belt A and Belt Elevator C).

\begin{figure}[ht]
 \centering
 \includegraphics[width=1.0\columnwidth,keepaspectratio]{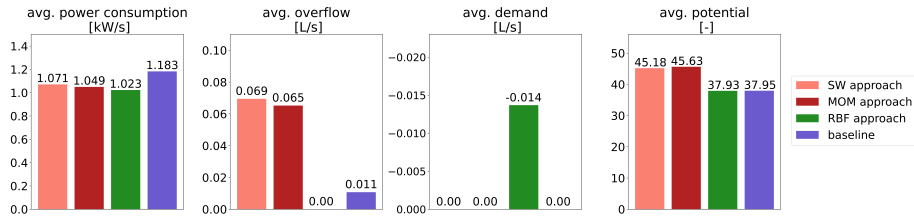}
\caption{Comparison between SbPGs and TL-SbPGs on the LS-BGS' performances.}
\label{fig:comparison_ls-bglp}
\end{figure}

In Fig.~\ref{fig:comparison_ls-bglp}, we present a performance comparison between the TL-SbPG methods and the baseline in the LS-BGS. The results show that the TL-SbPG approaches outperform the baselines in terms of power consumption and potential value. The TL-SbPG methods achieve reductions in power consumption of up to 13.53\%. Furthermore, TL-SbPGs using the SW and MOM approaches lead to a higher potential value compared to the baseline, where it is improved by approximately 20\%. Meanwhile, TL-SbPGs using the RBF model could not improve the system's performance.

Furthermore, we demonstrate that the number of players involved in transfer learning is not naturally limited. To this end, we conduct simulations with four different configurations of player involvement. Table~\ref{tab:ablation_num} presents the results for these varying player numbers using the MOM approach, which indicates that six players are the most effective in the LS-BGS. In the arbitrary environment experiments of the BGS, it is noted that involving four players in the 1-3-2//3-4 sequence is more advantageous than involving only two players. This suggests that the number of players is not restricted and can be increased as long as there are sufficient similar players available.
\begin{table}[ht]
\renewcommand{\arraystretch}{1.0}
\caption{Ablation study of the number of players involved in the transfer learning process in the LS-BGS.}
\label{tab:ablation_num}
\centering
\begin{tabular}{|ccccc|}
\hline
\multicolumn{1}{|c||}{\textbf{Algorithm}} & \multicolumn{1}{c|}{\textbf{\begin{tabular}[c]{@{}c@{}}Demand\\ (L/s)\end{tabular}}} & \multicolumn{1}{c|}{\textbf{\begin{tabular}[c]{@{}c@{}}Overflow\\ (L/s)\end{tabular}}} & \multicolumn{1}{c|}{\textbf{\begin{tabular}[c]{@{}c@{}}Power\\ (kW/s)\end{tabular}}} & \textbf{Potential} \\ \hline\hline
\multicolumn{1}{|c||}{Baseline} & \multicolumn{1}{c|}{0.000} & \multicolumn{1}{c|}{0.011} & \multicolumn{1}{c|}{1.183} & 37.951 \\ \hline\hline
\multicolumn{5}{|c|}{2P : P5-P10} \\ \hline\hline
\multicolumn{1}{|c||}{MOM} & \multicolumn{1}{c|}{0.000} & \multicolumn{1}{c|}{0.098} & \multicolumn{1}{c|}{1.087} & 42.703 \\ \hline\hline
\multicolumn{5}{|c|}{4P : P5-P10, P2-P6} \\ \hline\hline
\multicolumn{1}{|c||}{MOM} & \multicolumn{1}{c|}{0.000} & \multicolumn{1}{c|}{0.071} & \multicolumn{1}{c|}{1.052} & 44.940 \\ \hline\hline
\multicolumn{5}{|c|}{6P: P5-P10, P2-P6, P1-P4} \\ \hline\hline
\multicolumn{1}{|c||}{MOM} & \multicolumn{1}{c|}{0.000} & \multicolumn{1}{c|}{0.065} & \multicolumn{1}{c|}{1.049} & 45.634 \\ \hline\hline
\multicolumn{5}{|c|}{8P: P5-P10, P2-P6, P1-P4, P3-P8} \\ \hline\hline
\multicolumn{1}{|c||}{MOM} & \multicolumn{1}{c|}{-0.001} & \multicolumn{1}{c|}{0.101} & \multicolumn{1}{c|}{1.060} & 41.330 \\ \hline
\end{tabular}
\end{table}

From this ablation study, we observe that the distributed learning architecture imposes no limitation on the number of players in either the vanilla SbPG or TL-SbPG frameworks. The number of players involved in the transfer learning process is determined solely by the availability and similarity of players within the system. To be noted, the number of inputs to the learning algorithm is dependent on the number of neighbouring players. Consequently, the complexity of the system scales only with the number of neighbouring players, which is typically quite limited in production processes.

\begin{figure}[ht]
 \centering
 \includegraphics[width=0.85\columnwidth,keepaspectratio]{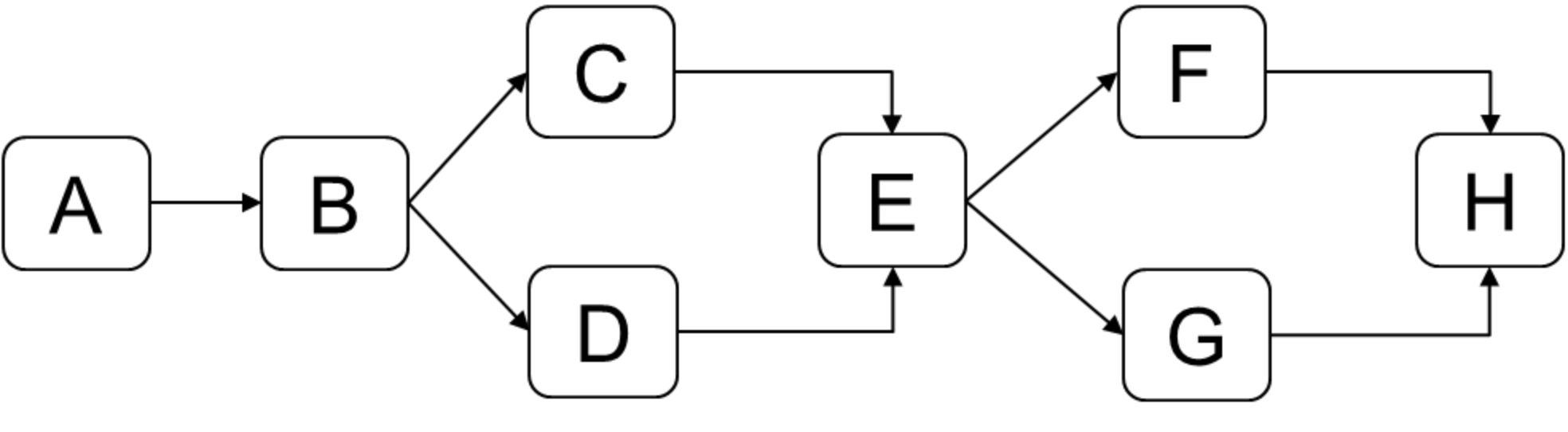}
\caption{Schematic of the LS-BGS with serial-parallel processes.}
\label{fig:comparison_ls-bglp_sp}
\end{figure}

We then extend our experimental setup of the LS-BGS to include serial-parallel processes, as illustrated in Fig.~\ref{fig:comparison_ls-bglp_sp}. Similar to the plug-and-play functionalities in the BGS, we can reuse the trained policies from the default sequential setup and retrain them with an additional three training episodes. We also maintain the same number of players involved in the transfer learning process, as determined from the previous ablation study. The results, presented in Table~\ref{tab:ablation_lsbgs}, indicate that the LS-BGS consistently outperforms the vanilla SbPG, while the RBF model does not significantly enhance the system's performance compared to the SW and MOM approaches. These findings demonstrate that TL-SbPGs exhibit robustness and scalability in more complex and varied environments, further validating our proof of convergence.
\begin{table}[ht]
\renewcommand{\arraystretch}{1.0}
\caption{Comparison between SbPGs and TL-SbPGs on the LS-BGS' performances with serial-parallel processes.}
\label{tab:ablation_lsbgs}
\centering
\begin{tabular}{|c||c|c|c|c|}
\hline
\textbf{Algorithm} & \textbf{\begin{tabular}[c]{@{}c@{}}Demand\\ (L/s)\end{tabular}} & \textbf{\begin{tabular}[c]{@{}c@{}}Overflow\\ (L/s)\end{tabular}} & \textbf{\begin{tabular}[c]{@{}c@{}}Power\\ (kW/s)\end{tabular}} & \textbf{Potential} \\ \hline\hline
Baseline & -0.001 & 0.169 & 1.479 & 35.406 \\ \hline
SW & 0.000 & 0.070 & 1.285 & 38.957 \\ \hline
MOM & 0.000 & 0.066 & 1.243 & 39.035 \\ \hline
RBF & -0.009 & 0.187 & 1.370 & 35.980 \\ \hline
\end{tabular}
\end{table}

A paired t-test was conducted to evaluate the performance of TL-SbPGs against SbPGs across six experiments in the LS-BGS, involving both sequential and serial-parallel processes with varied TL-SbPG variances. The analysis revealed a statistically significant performance advantage for TL-SbPG, with a mean utility improvement of 2.27 ($t(5) = 3.77, p = 0.0133$). These results indicate once again that TL-SbPGs offers superiority across different LS-BGS process structures.

After discussing the advantages of our proposed approaches in two experimental setups, we outline the limitations identified through our findings. The experiments revealed three limitations, although these do not diminish the overall contributions of the proposed TL-SbPGs approach. First, the method requires a digital representation of the system to train the players' policies due to the randomness during exploration, which makes it unsafe to train directly on real systems. This limitation can be addressed using a model-based GT approach, as suggested in~\citep{Yuwono2023a, Yuwono2023b}. Second, the SbPG utilizes best-response learning, which relies on ad-hoc random sampling for exploration. During periods of high exploration, we disabled the transfer learning mechanism. If a smoother sampling method were utilized for exploration, transfer learning could potentially be applied from the start, which leads to faster learning. Third, training the systems requires a large number of system interactions, which can be costly. This issue can also be mitigated through the model-based GT.

\section{Conclusions}\label{sec:conclusion}

We have presented a novel approach for transfer learning in SbPGs applied to distributed manufacturing systems, namely TL-SbPGs. Our contributions include proposing various transfer learning approaches for two different settings, where (a) the similarity between players is given beforehand and (b) the similarity has to be inferred during training. The proposed TL-SbPG approaches consist of the SW, MOM, and RBF model approaches. Through our experiments, we observed that the players were able to effectively transfer important knowledge to each other during training and utilize this knowledge to optimize their policies.

The TL-SbPG methods were implemented and evaluated in a modular production unit and a larger scale of the production unit with a multi-objective optimization scenario, where the experimental results demonstrate the effectiveness and promising nature of the TL-SbPG approaches compared to the baseline. We observed significant performance improvements, particularly in terms of power consumption optimization and the overall achievement of global objectives. In conclusion, our paper provides valuable insights and contributions to the field of transfer learning in SbPG for distributed manufacturing systems.

In future research, we aim to implement the proposed transfer learning methods across various game structures, including distributed Stackelberg strategies in SbPGs~\citep{Yuwono2024}. Additionally, we plan to integrate TL-SbPGs into the model-based GT domain, which enables learning within virtual environments. This would reduce the number of system interactions required and eliminate the need for a digital representation of the system. Additionally, we aim to enhance the transfer learning methods by developing a simpler approach that requires less communication between players to avoid noise and accelerate the transfer learning processes.

\section*{Funding}
This research did not receive specific funding from any grant provided by the public, commercial, or not-for-profit sectors.

\bibliographystyle{elsarticle-num}
\bibliography{sample}

\end{document}